\documentclass[10pt,twocolumn,letterpaper]{article}

\usepackage{wacv}
\usepackage{times}
\usepackage{epsfig}
\usepackage{graphicx}
\usepackage{amsmath}
\usepackage{amssymb}
\usepackage{float}
\usepackage{booktabs}

\makeatletter
\@namedef{ver@everyshi.sty}{}
\makeatother
\usepackage{tikz}
\usetikzlibrary{spy}
\usepackage[accsupp]{axessibility}  

\usepackage{comment}
\usepackage{color}
\usepackage{cite}
\usepackage{caption}
\usepackage{subcaption}
\usepackage{multirow}
\usepackage{bbold}
\usepackage{algorithm}
\usepackage[noend]{algpseudocode}

\def\wacvPaperID{973}
\wacvalgorithmstrack

\wacvfinalcopy

\ifwacvfinal
\usepackage[breaklinks=true,bookmarks=false]{hyperref}
\else
\usepackage[pagebackref=true,breaklinks=true,colorlinks,bookmarks=false]{hyperref}
\fi

\makeatletter
\newcommand{\multiline}[1]{
	\begin{tabularx}{\dimexpr\linewidth-\ALG@thistlm}[t]{@{}X@{}}
		#1
	\end{tabularx}
}
\makeatother

\algblockdefx[FstStep]{StartFstStep}{EndFstStep} 
{\textbf{First step}}%
{End of first step}
\algnotext[FstStep]{EndFstStep}	

\algblockdefx[SndStep]{StartSndStep}{EndSndStep} 
{\textbf{Second step}}%
{End of second step}
\algnotext[SndStep]{EndSndStep}

\newcommand{\prox}{\ensuremath{\operatorname{prox}}}
\newcommand{\Mod}[1]{\,(\mathrm{mod}\ #1)} 
\newcommand{\SecLinPnpADMM}{Section~1 of the Supplementary Material}
\newcommand{\SecClosedForm}{Section~2 of the Supplementary Material}

\newcommand{\Niters}{\ensuremath{N}}

\usepackage{pifont}
\newcommand{\cmark}{\ding{51}}%
\newcommand{\xmark}{\ding{55}}%

\begin{document}

\title{Deep Model-Based Super-Resolution with Non-uniform Blur}

\author{Charles Laroche\\
GoPro \& MAP5 \\
{\tt\small charles.laroche@u-paris.fr}
\and
Andrés Almansa\\
CNRS \& Université Paris Cité\\
{\tt\small andres.almansa@parisdescartes.fr}
\and
Matias Tassano\\
Meta Inc.\textsuperscript{*}\\
{\tt\small mtassano@meta.com}
}

\maketitle
\thispagestyle{empty}

\let\thefootnote\relax\footnotetext{\textsuperscript{*}Work mostly done while Matias was at GoPro France.}

\begin{abstract}

We propose a state-of-the-art method for super-resolution with non-uniform blur. Single-image super-resolution methods seek to restore a high-resolution image from blurred, subsampled, and noisy measurements. Despite their impressive performance, existing techniques usually assume a uniform blur kernel. Hence, these techniques do not generalize well to the more general case of non-uniform blur. Instead, in this paper, we address the more realistic and computationally challenging case of spatially-varying blur. To this end, we first propose a fast deep plug-and-play algorithm, based on linearized ADMM splitting techniques, which can solve the super-resolution problem with spatially-varying blur. Second, we unfold our iterative algorithm into a single network and train it end-to-end. In this way, we overcome the intricacy of manually tuning the parameters involved in the optimization scheme. Our algorithm presents remarkable performance and generalizes well after a single training to a large family of spatially-varying blur kernels, noise levels and scale factors.

\end{abstract}

\section{Introduction}

%
Single image super-resolution (SISR) methods aim to up-sample a blurred, noisy and possibly aliased low-resolution (LR) image into a high-resolution (HR) one. In other words, the goal of SISR is to enlarge an image by a given scale factor $s>1$ in a way that makes fine details more clearly visible.
%
The problem is ill-posed since there exist many ways to up-sample each low-resolution pixel. In order to further constrain the solution, a prior is usually imposed on the reconstructed HR output via a regularizer.
%
Early Bayesian and variational approaches to the SISR problem used Tikhonov \cite{milanfar2011SRbook,vsorel2017towards}, TV \cite{Malgouyres_2002,Almansa2004}, wavelet-$\ell_1$ \cite{escande_sparse_2013}, non-local \cite{Protter_2009}, or patch-recurrence \cite{Michaeli2013,Michaeli2014} regularization schemes, or adaptive filtering techniques~\cite{Romano2017} to impose a reasonable prior on the HR solution.
%
But classical regularization schemes have shown their limits. In order to cope with real world SISR problems, which are more ill-posed (higher noise levels, larger zoom factors, larger and more complex blur kernels) recent methods have turned to more powerful deep-learning-based regularizers, regressors or (conditional) generative models.
And they succeeded remarkably, producing extremely high-quality results for very large ($\times 16$) scale factors, as long as blur is uniform and small \cite{Saharia2021}.

\begin{figure}[btp]
\centering
\resizebox{\columnwidth}{!}{%
\begin{tabular}{cccc}
 LR & IKC & USRNet & Ours \\
\includegraphics[width=0.2\linewidth]{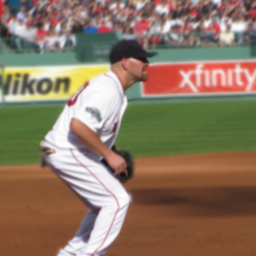} &
\includegraphics[width=0.2\linewidth]{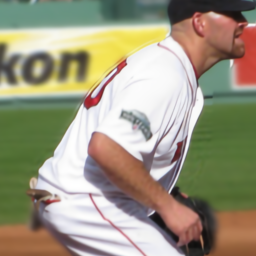} &
\includegraphics[width=0.2\linewidth]{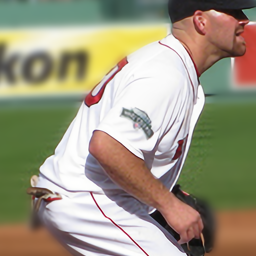} & 
\includegraphics[width=0.2\linewidth]{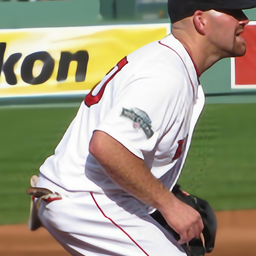}
\end{tabular}%
}
\caption{Super-resolution with scale factor 2 in the presence of spatially-varying blur. The foreground is not blurred while the background is blurred using isotropic Gaussian kernel.}
\label{fig:first_fig}
\end{figure}
%
The focus of this paper is on more realistic cases where blur kernels are non-uniform and much larger and complex, due mainly to motion blur and defocus blur~\cite{whyte10, 6583957}. Such degradations are very common in action cameras where the camera shake leads to spatially-varying motion blur or in microscopy where the lens blur cannot be assumed to be uniform. In such setting, doing super-resolution and deblurring in two steps is sub-optimal since it suffers from error accumulation. Also, the two steps approach does not exploit the correlation between the two tasks. Those observations raise the need for a super-resolution model robust to spatially-varying blur. This particular case received much less attention in the recent deep-learning-based SISR literature.
Among recent works, BlindSR~\cite{BlindSR} can handle non-uniform blur, but it does so only for relatively small and isotropic blur kernels which are quite far from real-world examples.
Other models such as USRNet~\cite{DUN}, can handle larger, anisotropic motion blur kernels, but they fail to generalize to spatially-varying blur.
This paper brings together these two characteristics to propose the first deep-learning based SISR method that can deal with both \emph{spatially-varying} and \emph{highly anisotropic, complex} blur kernels.

Like in~\cite{DUN}, our architecture is an unfolded version of an iterative optimization algorithm that solves the underlying posterior maximization problem. As demonstrated in~\cite{DUN}, this kind of model-based architecture provides a remarkable ability to generalize to a large family of blur kernels.
In order to allow our architecture to deal with spatially-varying blur, while keeping computational complexity low, we derive the unfolding from a linearized version of the ADMM algorithm.

%
The rest of the paper is organized as follows:
In Section~\ref{sec:related-work} we review the recent SISR literature with an emphasis in their support for spatially-varying and highly-anisotropic blur kernels. Figure~\ref{tab:model_tasks} summarizes our review which is later refined in Section~\ref{sec:compared-methods}.
Section~\ref{sec:model} introduces our architecture, its relationship to deep Plug \& Play, and linearized ADMM algorithms and provides details on how our end-to-end architecture has been trained.
The extensive experimental evaluation in  Section~\ref{sec:experiments} shows that our model significantly improves state-of-the-art performance on super-resolution with non-stationary blur, and that it can easily generalize to various non-uniform blur kernels, upscaling factors, and noise levels which is interesting for real-world applications. Training code and pre-trained model are available at: \url{https://github.com/claroche-r/DMBSR}

\section{Related Work}\label{sec:related-work}

\subsection{Learning Based Super-Resolution}
\label{sec:rel-cnn}
Several deep learning-based methods have been proposed to approach the SISR problem. SRCNN~\cite{SRCNN} is among the first models of this type. They employed a CNN to learn the mapping between an LR image and its HR version. Other methods used a similar approach but modified the architectures or losses~\cite{Lim2017, vdsr, recur_sr, persist_sr, atten_sr}.
ESRGAN~\cite{esrgan} introduced a GAN based loss to reconstruct high-frequency details along with an architecture based on the Residual in Residual Dense Block (RRDB). ESRGAN generates sharp and highly realistic super-resolution on synthetic data. However, it struggles to generalize on real images, as their training dataset is built using bicubic downsampling. To overcome this limitation, BSRGAN~\cite{BSRGAN} proposes to retrain ESRGAN on a more realistic degradation pipeline.
Unlike previous SISR methods which disregard the blur kernel, SRMD~\cite{zhang2018learning}
proposes to give the blur kernel information as an additional input to the network. This method then belongs to the family of so-called \emph{non-blind} SISR methods.
\emph{Blind} methods such as~\cite{KernelGAN, IKC, BlindSR} tackle the issue of kernel estimation for super-resolution using an internal-GAN, iterative kernel refinement or a dedicated discriminator. In~\cite{ZSSR, DualSR}, the authors address the issue of generalization using an image-specific super-resolver trained using cyclic loss on intrinsic patches of the low-resolution image.
Recently, SwinIR~\cite{SwinIR} proposes a swin transformer~\cite{swin} based architecture that achieves state-of-the-art results while heavily reducing the number of parameters of the network.
\begin{figure}[btp]
    \centering
    \begin{subfigure}{0.50\textwidth}
    \includegraphics[width=0.9\textwidth]{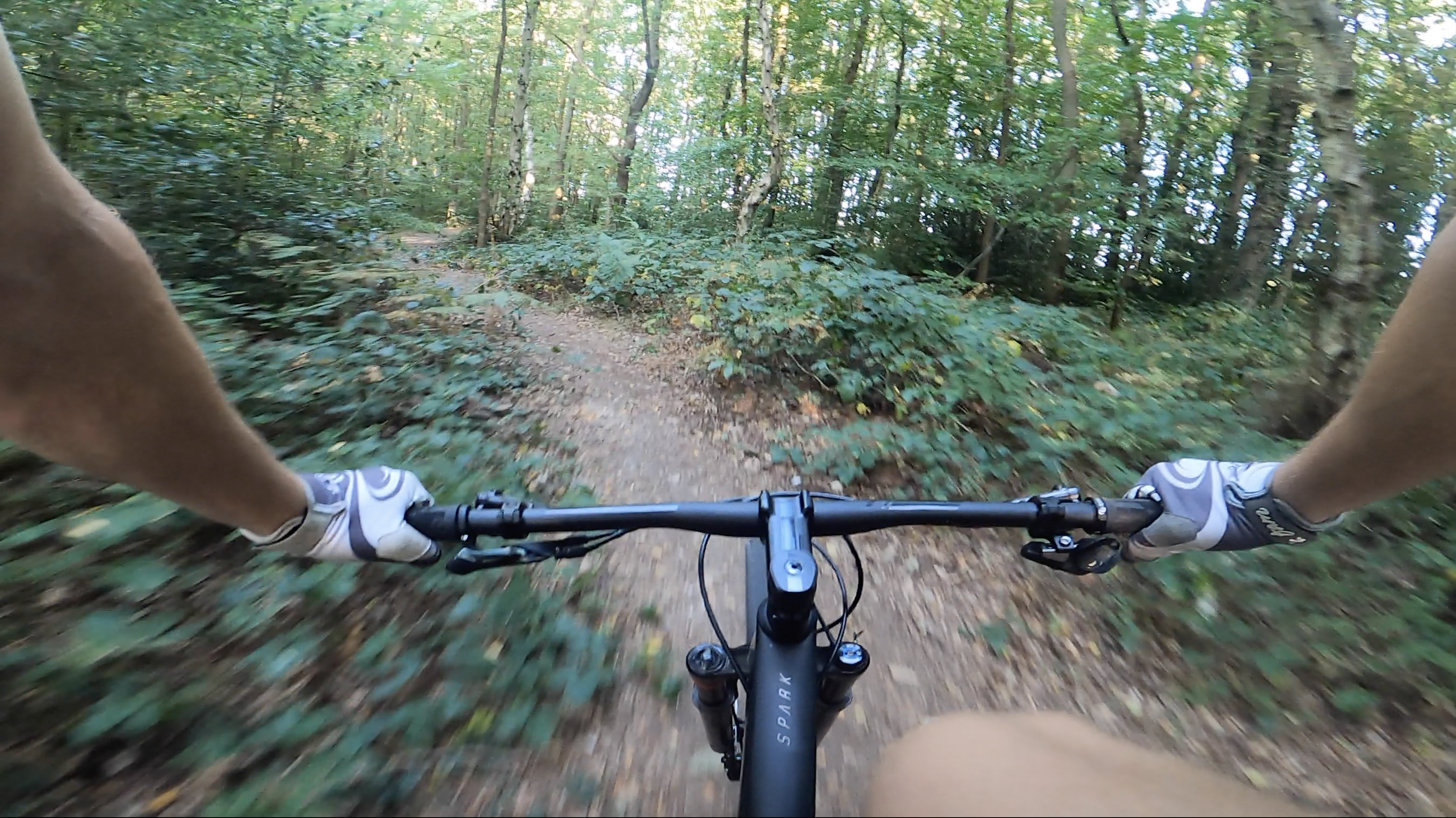}
    \caption{Example of spatially-varying blur.}
    \label{fig:motion_blur}
    \end{subfigure}
    \begin{subfigure}{0.40\textwidth}
        \resizebox{\columnwidth}{!}{%
        \begin{tabular}{|c|c|c|c|c|c|}
         \hline
         Article & SR & UB & SVB & MB & Blind \\
         \hline
         Bicubic & \cmark & \xmark & \xmark & \xmark & \cmark \\
         ESRGAN~\cite{esrgan} & \cmark & \xmark & \xmark & \xmark & \cmark \\
         BSRGAN~\cite{BSRGAN} &\cmark & \cmark & \xmark & \xmark & \cmark \\
         SwinIR~\cite{SwinIR} & \cmark & \cmark & \xmark & \xmark & \cmark \\
         IKC~\cite{IKC} & \cmark & \cmark & \xmark & \xmark & \cmark \\
         BlindSR~\cite{BlindSR} & \cmark & \cmark & \cmark & \xmark & \cmark \\
         ZSSR~\cite{ZSSR} & \cmark & \cmark & \xmark & \xmark & \xmark \\
         DualSR~\cite{DualSR} & \cmark & \cmark & \xmark & \xmark & \cmark \\
         USRNet~\cite{DUN} & \cmark & \cmark & \xmark & \cmark & \xmark \\
         \hline \hline
         Architecture & SR & UB & SVB & MB & Blind \\
         \hline
         RRDB~\cite{esrgan} & \cmark & \cmark & (\cmark) & (\cmark) & \cmark \\
         SwinIR~\cite{SwinIR} & \cmark & \cmark & (\cmark) & (\cmark) & \cmark \\
         SFTMD + PCA~\cite{IKC} & \cmark & \cmark & \cmark & (\xmark) & \xmark \\
         BlindSR~\cite{BlindSR} & \cmark & \cmark & \cmark & (\xmark) & \xmark \\
         Ours & \cmark & \cmark & \cmark & \cmark & \xmark \\
         \hline
        \end{tabular}%
        }
    \caption{Method comparison table.}
    \label{tab:model_tasks}
    \end{subfigure}
    \caption{
    \emph{(a)} Background objects are moving with respect to the camera, so they appear blurry, whereas foreground objects are sharp.
    \emph{(b)} Restoration problems that \emph{are addressed by previous \underline{articles}} or that \emph{can be potentially solved by available \underline{architectures}:} SR=single image Super-Resolution; UB = Uniform Blur; SVB = Spatially Varying Blur; MB = Motion Blur. Brackets stand for potential use case of the architecture that have not been tested in the literature to our knowledge.
    }
\end{figure}

\subsection{Deep Plug-and-Play}

Deep plug-and-play methods can be traced back to~\cite{venkatakrishnan2013} where the image restoration problem is solved using ADMM optimization by decoupling the data and regularization terms. Then, they use a denoiser to solve the regularization sub-problem. This idea has been extended to other optimization schemes such as primal-dual methods~\cite{dpnp_dualprimal, Meinhardt2017} or fast iterative shrinkage algorithm~\cite{dpnp_fista}. A large diversity of denoisers have been used for the regularization. Among them, BM3D has been used the most~\cite{dpnp_dualprimal, dpnp_bm3d, dpnp_fista}, but more recently deep CNN based denoisers have become very popular~\cite{Meinhardt2017, dpnp_cnn_2}.~\cite{zhang2021plug} provides an analysis of the efficiency of the different deep denoisers for different image restoration tasks. Deep plug-and-play methods can be used to solve a large variety of image resoration tasks such as Gaussian denoising~\cite{dpnp_gauss}, image deblurring~\cite{dpnp_deblur} or super-resolution~\cite{dpnp_sr}. Theoretical aspects of deep plug-and-play algorithms have also been studied using bounded denoisers assumptions~\cite{dpnp_theory_bounded} or more recently using denoisers whose residual operators are Lipschitz-continuous~\cite{pmlr-v97-ryu19a, Laumont2022pnpsgd}. More recently,~\cite{DUN} built a deep unfolding network called USRNet for super-resolution using deep plug-and-play optimization.

\subsection{Spatially-Varying Blur}

Removing uniform blur is a well-studied problem. Classical methods design natural image priors such as $\ell_1$~\cite{l1deblur},  $\ell_2$~\cite{l2deblur} or hyper-Laplacian~\cite{NIPS2009_3dd48ab3}.
CNN learning-based approaches usually build coarse-to-fine deep learning architectures such as~\cite{7274732}, where CNN blocks simulate iterative optimization, or~\cite{8578951} which deblurs the image using a scale-recurrent network.
The task becomes much more complex when the blur varies spatially. Early approaches decompose the spatially-varying blur into a finite basis of spatially-uniform blur kernels and their respective spatially-varying weights~\cite{oleary}. This approach drastically reduces the dimension of the blur operator and makes it computable in a reasonable amount of time using Fourier transform. \cite{eff} build an alternative model designed for faster computation and apply it to deconvolution of spatially-varying blur. More recently,~\cite{escande_diag_2013, escande_sparse_2013, escande2018} approximate the spatially-varying blur operator in the wavelet basis by a diagonal or sparse operator. Their decomposition allows very efficient computation of the blur operator and its transpose since the structure of the operator allows GPU parallelization. Other approaches such as~\cite{eboli2020end2end} use HQS splitting to decouple the prior and data term. The data step is computed using an approximation of the inverse blur and the prior step is solved using CNN priors. 
Jointly solving the non-uniform deblurring and upsampling (super-resolution) problem is a much more challenging task that has been much less studied~\cite{BlindSR}. Most spatially-variant deblurring methods require the blur operator to be known. Estimating the non-uniform blur kernel has been tackled for several applications such as defocus~\cite{Ikoma2021dfd}, lens abberation~\cite{Delbracio2011psf} super-resolution~\cite{MaNet} and motion blur~\cite{carbajal2021nonuniform}.

\section{Model}\label{sec:model}
\subsection{Problem Formulation}

The standard model for single-image super-resolution with multiple degradations usually assumes that the low-resolution image is a blurry, noisy and subsampled version of a given high-resolution image,
\begin{equation}
    y = (x \circledast k)\downarrow_s + \epsilon \ \ \text{with} \ \ \epsilon \sim \mathcal{N}(0, \sigma^2),
    \label{equ:degr_model_class}
\end{equation}
with $x$ the high-resolution image, $y$ its low-resolution version, $k$ the blur kernel, $\downarrow_s$ the subsampling operator with scale factor s, and $\epsilon$ the noise. This formulation assumes that the blur kernel is uniform all over the image which makes the computation of the low-resolution image fast using convolution or fast Fourier transform. This assumption is not always realistic since camera or object motion will often result in non-uniform blur as illustrated in Figure~\ref{fig:motion_blur}. In this example, background objects are moving with respect to the camera, so they appear blurry, whereas foreground objects are sharp. Spatially-varying blur can also appear when the objects are out-of-focus. In this case, the blur is closely related to the depth of field.
Taking into account spatially-varying blur, the degradation model in Equation~\eqref{equ:degr_model_class} replaces the convolution operator with a more general blur operator that varies across the pixels
\begin{equation}
\label{eq:degr_model}
    y = (Hx)\downarrow_s + \epsilon \ \ \text{with} \ \ \epsilon \sim \mathcal{N}(0, \sigma^2),
\end{equation}
where $Hx$ corresponds to the non-uniform blur operator applied to image $x$ (flattened as a column vector). Working with unconstrained $H$ leads to computationally expensive operations. In our work, the only restriction we make on $H$ is that $Hx$ and $H^Tx$ must be computed in a reasonable amount of time. A basic example for such a use case is the O'Leary model~\cite{oleary} where $H$ is decomposed as a linear combination
\begin{equation}
\label{equ:oleary_1}
    H=\sum\limits_{i=1}^P{U_i K_i}
\end{equation}
of uniform blur (convolution) operators $K_i$ with spatially varying mixing coefficients, \emph{i.e.} diagonal matrices $U_i$ such that $\sum\limits_{i=1}^P{U_i} = Id$, $U_i \geq 0$. 
This model provides a convenient approximation of a spatially-varying blur operator, by reducing both the memory and computing resources required to store and compute this operator. In practice, $K_i$ may represent a basis of different kinds of (defocus or motion) blur, and the $U_i$ can represent object segmentation masks or sets of pixels with similar blur kernels.

\subsection{Deep Plug-and-Play}
\label{subsec:deepPnP}

Model-based variational or Bayesian methods usually solve the SISR problem in Equation~\eqref{eq:degr_model} by imposing a prior with density $p(x) \propto e^{-\lambda \phi(x)}$ to the HR image $x$ (common choices for $\Phi$ are Total Variation or $\ell_1$ norm of wavelet coefficients).
Then the maximization of the posterior density $p(x|y) \propto p(y|x)p(x)$ leads to the following optimization problem to compute the MAP estimator:
\begin{align}\label{eq:MAP}
     x^* & = \arg\min_x \frac{1}{2\sigma^2}\| (Hx)\downarrow_s - y\|_2^2 + \lambda \Phi(x) \\
     & = \arg\min_x g(Hx) + \lambda \Phi(x). \label{equ:define_g} 
\end{align}
This family of optimization problems is often solved using iterative alternate minimization schemes, like ADMM \cite{Boyd2010a}, which leads to iterating the following steps for $k=0,\dots,\Niters$:
\begin{align}
\label{equ:x-update}
& x_{k+1} = \prox_{(\lambda/\mu)\Phi(.)}(v_k - u_k)
= \mathcal{P}_{\sqrt{\lambda/\mu}}(v_k-u_k) \\
\label{equ:proxH}
    & v_{k+1} = \prox_{(1/\mu)g(H.)}(x_{k+1}+u_k) \\
\label{equ:u-update}
     & u_{k+1} = u_k + (x_{k+1} - v_{k+1}).
\end{align}
where $\prox_{\lambda f}$ is the proximal operator of $\lambda f$ defined by $\prox_{\lambda f}(v) = \arg\min_x {f(x) + (1/2\lambda)\|x-v\|_2^2}$ and $g$ is defined in~\ref{equ:define_g}.
For convex $\Phi$, this is known to converge to the solution of~Equation~\eqref{eq:MAP} as $k\to\infty$.
Deep plug-and-play methods use more sophisticated  (possibly non-convex) learned regularizers by simply replacing the regularization step in Equation~\eqref{equ:x-update}, by a CNN denoiser $\mathcal{P}_\beta$ which was pretrained to remove zero-mean Gaussian noise with variance $\beta^2$. The convergence of the iterative plug-and-play ADMM scheme still holds in this non-convex case for careful choices of the denoiser and the hyperparameters $(\lambda,\mu)$ \cite{pmlr-v97-ryu19a}.

In the case of a uniform blur, the $v$-update can be efficiently computed using the fast Fourier transform~\cite{DUN, FastSR} since the operator $H$ is diagonal in the Fourier basis. 
However, this is no longer the case for spatially-varying blur. Even for simpler use cases such as the O'Leary model from Equation~\eqref{equ:oleary_1} solving the subproblem \eqref{equ:proxH} can be very computationally expensive.
To avoid this, linearized ADMM \cite[sec 4.4.2]{prox_algo} (see also \cite{zhouchen2011, stoch_ADMM, Esser2010, Zhang2011}) substitutes the splitting variable $v=x$ by $z=Hx$, and introduces the linear approximation $\|Hx-z_k\|^2 \approx \mu( H^THx_k-H^Tz_k)^Tx + \frac{\rho}{2}\|x - x_k\|^2$ in the augmented Lagrangian, in order to avoid the need to invert $H$. This approximation corresponds to a linearization of the regularization $\|Hx-z\|_2^2$ with an extra regularization $\|x-x_k\|_2^2$ that enforces $x_k$ to be close the the linearization point of application. As a consequence, Equations~\eqref{equ:x-update}~to~\eqref{equ:u-update} are rewritten as follows:
\begin{align}
\label{equ:x-update_2}
x_{k+1} & = \prox_{(\lambda/\rho)\Phi}(x_k - (\mu / \rho)H^T(Hx_k - z_k + u_k))  \nonumber \\
& = \mathcal{P}_{\sqrt{\lambda/ \rho}}(x_k - (\mu / \rho)H^T(Hx_k - z_k + u_k)) \\
\label{equ:z-update_2}
z_{k+1} & = \prox_{(1/\mu)g(.)}(Hx_{k+1} + u_k) \\
\label{equ:u-update_2}
u_{k+1} & = u_k + Hx_{k+1} - z_{k+1},
\end{align}
with $\mu, \rho$ hyper-parameters of the optimization scheme. \newline
\newline
Now Equation \eqref{equ:z-update_2} can be easily computed in closed form since it does not require to invert a matrix involving $H$ anymore (see \SecClosedForm\ for more information). We have:
\begin{equation}
\label{equ:data-term}
    z_{k+1}(i,j) = \frac{\left((y\uparrow_s) + \sigma^2 \mu (Hx_{k+1} + u_k)\right)(i,j)}{\sigma^2 \mu  + \delta_{i \equiv 0 \Mod{s}}\delta_{j \equiv 0 \Mod{s}}},
\end{equation}
where $(.)\uparrow_s$ corresponds to the zero-padding up-sampling with scale factor $s$ and $\delta_{i \equiv 0 \Mod{s}}\delta_{j \equiv 0 \Mod{s}}$
is the indicator function that is equal to 1 on the pixels divided by the scale factor and 0 otherwise. The whole deep plug-and-play iterative program is summarized in Algorithm~\ref{alg:pnp-sr}.
\begin{algorithm}[btp]
\caption{Deep Plug-and-Play Linearized ADMM algorithm
\\
{\small
Solves $x = \arg\min_x \frac{1}{\sigma^2}\|(Hx)\downarrow_s - y \|^2 + \lambda \Phi(x)$ \\
using denoiser $\mathcal{P}_{\beta} = \prox_{\beta^2 \Phi}$ as an implicit regularizer.
}
}\label{alg:pnp-sr}
\begin{algorithmic}
\Require Measurements $y$, spatially varying kernel $H$, scale factor $s$, noise level $\sigma$, number of iterations $\Niters$, hyper-parameters $\lambda, \mu, \rho$
\Ensure super-resolved, deconvolved and denoised image $x_\Niters$
\State $x_0 \gets y\uparrow_s$
\State $z_0 \gets Hx_0$
\State $u_0 \gets 0$
\For{$k \in [0, \Niters-1]$}
    \State $x_{k+1} = \mathcal{P}_{\sqrt{\lambda/ \rho}}(x_k - (\mu / \rho) H^T(Hx_k - z_k + u_k))$
    \State $ z_{k+1}(i,j) = \frac{\left((y\uparrow_s) + \sigma^2\mu (Hx_{k+1} + u_k)\right)(i,j)}{\sigma^2 \mu + \delta_{i \equiv 0 \Mod{s}}\delta_{j \equiv 0 \Mod{s}}}$
    \State $u_{k+1} = u_k + Hx_{k+1} - z_{k+1}$
\EndFor
\end{algorithmic}
\end{algorithm}
The linearized ADMM algorithm in Equations~\eqref{equ:x-update_2} to \eqref{equ:u-update_2} was not yet studied in the Plug \& Play context, but recent results in~\cite{Liu2019} and \cite{Gribonval2011} suggest that it actually converges for careful choices of the parameters $\lambda, \mu, \rho$ (see \SecLinPnpADMM\ for details). These theoretical results motivated the Unfolded version of the linearized PnP-ADMM algorithm that we present in the next section, and is the basis of the experimental results in Section~\ref{sec:experiments}.

\subsection{Deep Unfolding Networks}

\begin{figure*}[btp]
    \centering
    \includegraphics[width=0.8\linewidth]{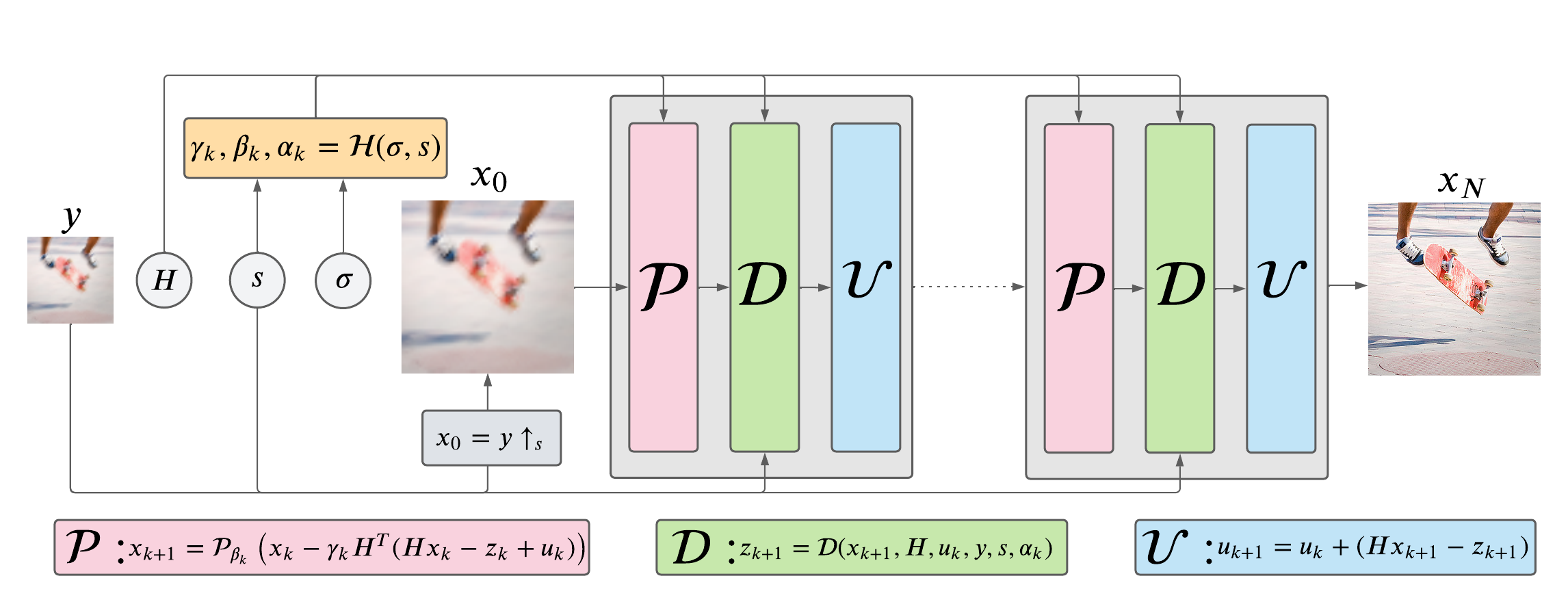}
    \caption{Model architecture, the low-resolution image is upsampled and alternately fed to the prior module $\mathcal{P}$, the data module $\mathcal{D}$ and the update module $\mathcal{U}$ during \Niters\ iterations}
    \label{fig:archi}
\end{figure*}
Deep plug-and-play methods achieve impressive performance on image restoration tasks. However, their efficiency strongly relies on the choice of their hyper-parameters. Finding correct values for the latter can be challenging. These methods also require a sufficient number of steps to properly converge, which is time-consuming. We improve the runtime and simplify the hyper-parameter tuning process by unfolding our algorithm into a deep learning architecture. This architecture is composed of a fixed and small number of iterations of the linearized ADMM algorithm and a MLP that automatically selects the hyper-parameters. The whole network is optimized using end-to-end training. Following the linearized ADMM formulation, the architecture alternates between a prior-enforcing step $\mathcal{P}$ corresponding to Equation~\eqref{equ:x-update_2}, a data-fitting step $\mathcal{D}$ (see Equation~\eqref{equ:z-update_2}) and finally an update block $\mathcal{U}$ (see Equation~\eqref{equ:u-update_2}). These blocks are respectively stacked $\Niters$ times corresponding to the number of iterations. The optimization process requires the hyper-parameter triplets $(\lambda, \mu, \rho)$ that are predicted by the hyper-parameters block $\mathcal{H}$ at each step. The resulting architecture of our deep-unfolding network is presented in Figure \ref{fig:archi}. Next, we present each block in detail.
\newline
\newline
The first block of our network is the prior module $\mathcal{P}$. As explained in Section~\ref{subsec:deepPnP}, Equation~\eqref{equ:x-update_2} corresponds to a denoising problem, which is approximated by a CNN denoiser. Based on the work in~\cite{DUN}, we use a ResUNet~\cite{ResUNet} architecture with the denoising level as an extra input for $\mathcal{P}$. All the parameters of the ResUNet are learned during the end-to-end training process. Retraining the parameters of the network helps to obtain the best quality results for the given number of iterations. The $x$-update is finally expressed as:
\begin{equation}
    x_{k+1} = \mathcal{P}_{\beta_k}\left(x_k - \gamma_k H^T (H x_k - z_k + u_k)\right),
\end{equation}
with $\beta_k = \sqrt{\lambda_k/\rho_k}$ and $\gamma_k = \mu_k / \rho_k$. The splitting algorithm introduces the quantity $x_k - \gamma_k H^T (H x_{k+1} - z_k + u_k)$. This quantity can be interpreted as a deblurring gradient descent step on the current clean estimate $x_k$. The $x$-update combines the deblurring and denoising operations.
\newline
\newline
The data-term module $\mathcal{D}$ computes the proximal operator of $z \mapsto \frac{1}{2\sigma^2}\|z\downarrow_s - y\|_2^2$ at $z=H x_{k+1} + u_k$. Following Equation~\eqref{equ:data-term}, we re-write our data-term as:
\begin{equation}
    z_{k+1} = \mathcal{D}(x_{k+1}, H, u_k, y, s, \alpha_k),
\end{equation}
with $\alpha_k = \sigma^2 \mu_k$. The data term will ensure that our current estimate of the sharp image is consistent with the degraded input. It also acts as a mechanism of injection of the degraded image $y$ through the iterations.
\newline
\newline
The update module $\mathcal{U}$ updates the dual variable (or Lagrange multiplier) $u$ of the ADMM algorithm. This block does not have trainable parameters. We decided to integrate this step into the architecture to be consistent with the ADMM formulation.
\newline
\newline
Finally, the $(\gamma_k, \beta_k, \alpha_k)$ hyper-parameters of the plug-and-play model are predicted as a function of noise level $\sigma$ and scale factor $s$ by a neural network  $\mathcal{H}$. Indeed $\alpha_k = \sigma^2 \mu_k$ directly depends on $\sigma$ and $\beta_k = \sqrt{\lambda_k/\rho_k}$ depends on the regularization parameter $\lambda$ whose optimal value is usually affected by both $\sigma$ and $s$.
For the architecture of $\mathcal{H}$, we use 3 fully connected layers with ReLU activations. The dimension of the hidden layers is 64.

\subsection{Training}

The architecture is trained end-to-end using the L1 loss for 200 epochs. We start with a learning rate of 1e-4 and decrease it every 50 epochs by a factor 0.1. The ResUNet parameters were initialized by a pre-trained model that solves a Gaussian denoising problem. We found that doing so improves the stability of the model during training. We use \Niters=8 iterations in our unfolded architecture for the experiments. The network is trained using scale factors $s \in \{1,2,3,4\}$, noise levels $\sigma \in [0, 25]$ and spatially varying blur kernels composed various motion blurs and Gaussian blurs.

\section{Experiments}\label{sec:experiments}

\subsection{Data Generation}\label{sec:data-generation}

\begin{figure}
\centering
\begin{subfigure}{0.5\textwidth}
    \centering
    \includegraphics[width=.8\linewidth]{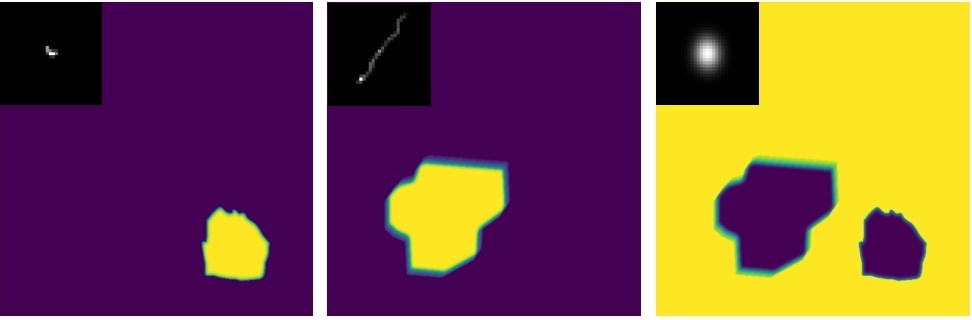}
    \caption{Object masks $U_i$ and kernels $K_i$}
\end{subfigure}
\begin{subfigure}{0.5\textwidth}
    \centering
    \includegraphics[width=.8\linewidth]{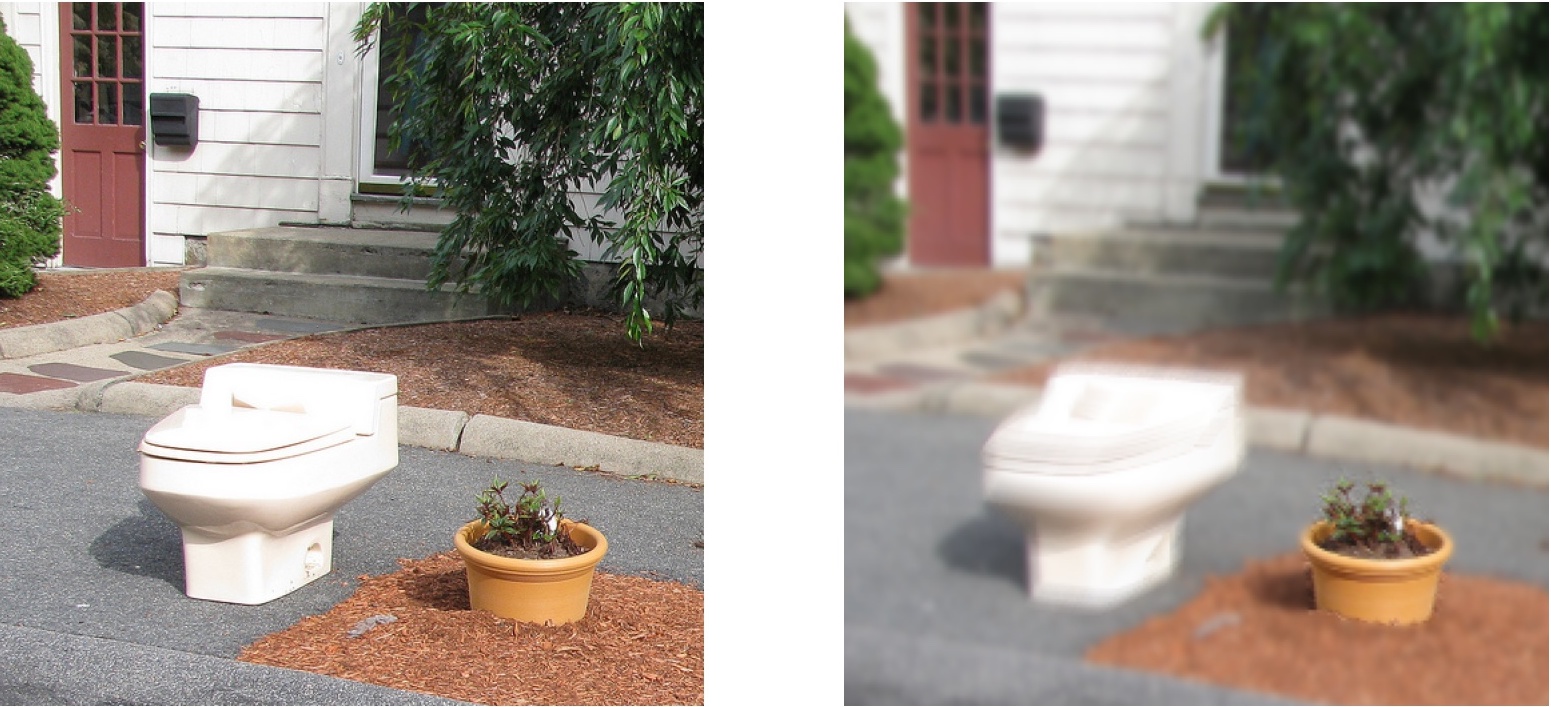}
    \caption{Generated pairs}
\end{subfigure}
\caption{Example of data generated by our pipeline}
\label{fig:data_pipe}
\end{figure}

Gathering real-world data with spatially-varying blur and their respective kernels is very complicated. Instead, we train our model using synthetic data. For this experiment, we adopted the O'Leary blur model from Equation \eqref{equ:oleary_1}. This blur decomposition covers a large variety of spatially-varying blurs ranging from motion blur to defocus blur. Figure \ref{fig:data_pipe} represents an example of synthetic blur obtained using this formulation.
For the training and testing, we used COCO dataset~\cite{coco}. COCO dataset is a well-known large-scale dataset for object detection, segmentation,and image captioning. It is composed of more than 200K images segmented for 80 object categories and 91 stuff categories representing 1.5 million object instances. We use the segmentation masks to blur the objects and the background. We use both Gaussian kernels and motion blur kernels. We build a database of motion blur kernels using~\cite{camera_shake}.
To ensure a smooth and realistic transition between the blurred areas, we blur the borders of the masks so that a mix between blurs occurs at the edges of the objects. We finally normalize the masks so that their sum is equal to 1 for each pixel. After blurring the image, we apply nearest neighbor downsampling with scale factor varying in $\{1,2,3,4\}$ and Gaussian blur with $\sigma \in [0,25]$.

Finally, our data generation pipeline implements the following degradation model:
\begin{equation}
\label{equ:oleary_2}
    y = (\sum\limits_{i=1}^P{U_i K_i x})\downarrow_s + \epsilon,
\end{equation}
with $x$ the clean image, $y$ its low-resolution version, $s$ the scale factor, and $\epsilon$ the noise. 

\subsection{Compared Methods}\label{sec:compared-methods}

We compare the proposed model to Bicubic upsampling (widely used baseline), RRDB~\cite{esrgan}, SwinIR~\cite{SwinIR}, SFTMD~\cite{IKC}, BlindSR~\cite{BlindSR} and USRNet~\cite{DUN}.

Few super-resolution models can generalize to non-uniform blur (see Table~\ref{tab:model_tasks}). We believe that the models listed above represent the current most pertinent solutions for such a setting. However, using the pre-trained weights from the source code of each model leads to poor performance on our testing dataset since they are trained using uniform blur kernels. In order to ensure a fair comparison with our model, we retrain all those architectures on our database. We use the MSE loss for the retraining of all models so that the PSNR is maximized. Next, we present in details how we use those architectures. 

RRDB and SwinIR described in section~\ref{sec:rel-cnn} are blind methods so we just retrain them on our dataset using the configuration given by the authors.

SFTMD is the non-blind architecture introduced in~\cite{IKC} that combines kernel encoding using Principal Components Analysis (PCA) and spatial features transforms (SFT)~\cite{sft} layers. The PCA-encoded blur kernel is fed to the SFT-based network along with the low-resolution image. In the case of spatially-varying blur, they encode the kernels at each pixel’s location and give the resulting spatially-varying map of encoded kernels to the network.

BlindSR proposes an alternative to the PCA for the encoding of the kernel. They use an MLP that is trained along with the super-resolution network. This allows the network to encode more complex kernels. We use the non-blind part of BlindSR that is composed of a backbone with convolutions, dense layers and residual connections with MLP encoding for the kernel.

We finally compare our architecture to USRNet, which is similar to our model but works only with uniform blur. Since we work with the O'Leary model \eqref{equ:oleary_1}, we can apply USRNet on each blurred mask $U_i$ with their corresponding uniform blur kernel $K_i$ and then reconstruct the results by summing the output of the model on each mask. Since we work with the classical USRNet, we do not retrain it and use the weights from the source code of the method.
\emph{i.e.}
$ \operatorname{USRNet}(y,H) \approx \sum_{i=1}^P U_i \operatorname{USRNet}(y,K_i)$

\subsection{Quantitative Results}
Table \ref{table:res} summarizes the PSNR, SSIM (structural similarity index) and LPIPS (learned perceptual image patch similarity) on the different testsets. The testsets are constructed from the COCO validation set and the degradation model of Equation \ref{equ:oleary_2}. Performance on Gaussian and motion blur are evaluated separately. We test the models on x2 and x4 super-resolution without additive noise.
\begin{table}
\begin{center}
\caption{Quantitative comparison on synthetic data. The displayed metrics correspond respectively to PSNR$\uparrow$, SSIM$\uparrow$ and LPIPS$\downarrow$. Best scores are displayed in red, second bests in blue.}
\resizebox{\columnwidth}{!}{%
\begin{tabular}{| c | c | c c c |}
\hline
Scale & Type & Model  & Gaussian blur & Motion blur \\
\hline\hline
\multirow{7}{5em}{x2} & \multirow{3}{5em}{Blind} &  Bicubic & (22.52, 0.60, 0.57) & (21.74, 0.62, 0.39)\\
& & RRDB~\cite{esrgan} & (23.38, 0.67, 0.41) & (23.11, 0.65, 0.36)\\
& & SwinIR~\cite{SwinIR} & (23.47, 0.67, 0.38) & (23.40, 0.67, 0.34)\\
\cline{2-5}
& \multirow{4}{5em}{Non-blind} & SFTMD~\cite{IKC} & (23.76, 0.69, 0.33) & (25.15, 0.74, 0.25) \\
& & BlindSR~\cite{BlindSR} & (\color{blue}{26.55}, \color{red}{0.79}, \color{red}{0.24}) & (\color{blue}{26.40}, \color{blue}{0.79}, 0.20)\\
& & USRNet~\cite{DUN} & (22.64, 0.74, 0.28) & (24.37, 0.75, \color{blue}{0.17})\\
& & Ours  & (\color{red}{26.59}, \color{blue}{0.78}, \color{blue}{0.26})&(\color{red}{28.20}, \color{red}{0.85}, \color{red}{0.11})\\
\hline
\multirow{7}{5em}{x4} & \multirow{3}{5em}{Blind} &  Bicubic & (21.61, 0.55, 0.60) & (20.48, 0.56, 0.57)\\
& & RRDB~\cite{esrgan} & (21.82, 0.57, 0.58) & (22.34, 0.60, 0.56)\\
& & SwinIR~\cite{SwinIR} & (23.01, 0.63, 0.44) & (22.70, 0.64, 0.44)\\
\cline{2-5}
& \multirow{4}{5em}{Non-blind} & SFTMD~\cite{IKC} & (23.12, 0.64, 0.41) & (23.97, 0.67, 0.38) \\
& & BlindSR~\cite{BlindSR} & (\color{blue}{25.11}, \color{blue}{0.72}, \color{black}{0.34}) & (24.54, 0.69, 0.35)\\
& & USRNet~\cite{DUN} & (24.08, \color{blue}{0.72}, \color{blue}{0.32})  & (\color{blue}{24.67}, \color{blue}{0.72}, \color{blue}{0.29})\\
& & Ours  & (\color{red}{25.37}, \color{red}{0.73}, \color{red}{0.31}) & (\color{red}{25.36}, \color{red}{0.73}, \color{red}{0.28})\\
\hline \end{tabular}
\label{table:res}
}
\end{center}
\end{table}
Firstly, non-blind models outperform blind ones by a large margin. The extra information about the degradation is well used by the networks. The blind transformers-based architecture of SwinIR is more efficient than the classical RRDB.
For the non-blind architectures, we can see the importance of how the blur operator information is given to the network. Specifically, the neural network encoding of the blur operator from BlindSR outmatches by far the PCA encoder from the IKC version of SFTMD. 
It highlights the fact that the PCA model is not complex enough to capture interesting features of the blur kernels. The BlindSR model performs well on the Gaussian testset but fails to generalize on motion blur. One reason for this is the fact that motion blurs are too complex to be encoded by PCA or a small MLP. The USRNet model reaches good SSIM and LPIPS whilst having low PSNR. This is mostly due to the fact that this model introduces artefacts which are not captured by SSIM or LPIPS. The poor performances of USRNet underline the fact that networks trained on uniform blur cannot naively generalize well to spatially-varying blur even on simple use cases. Finally, our model outperforms all the other methods by an average of 0.15dB for the Gaussian blur testsets and 1.2dB on the motion blur testset. Additionally, our algorithm outperform all other methods on SSIM and LPIPS, except for the x2 Gaussian blur case.
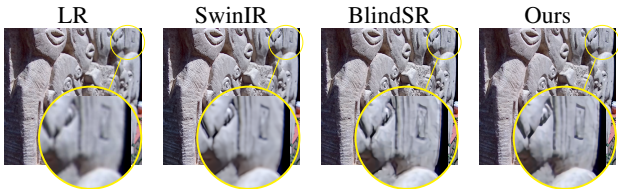
\begin{figure}[t]
\centering
\resizebox{\columnwidth}{!}{%
\begin{tikzpicture}[spy using outlines={circle,yellow,magnification=3,size=1.5cm, connect spies}]
\node {\includegraphics[height=2cm]{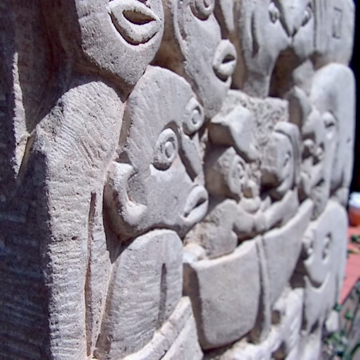}};
\spy on (0.8,0.8) in node [left] at (1,-0.6);
\draw (0, 1.2) node {LR};
\end{tikzpicture}
\begin{tikzpicture}[spy using outlines={circle,yellow,magnification=3,size=1.5cm, connect spies}]
\node {\includegraphics[height=2cm]{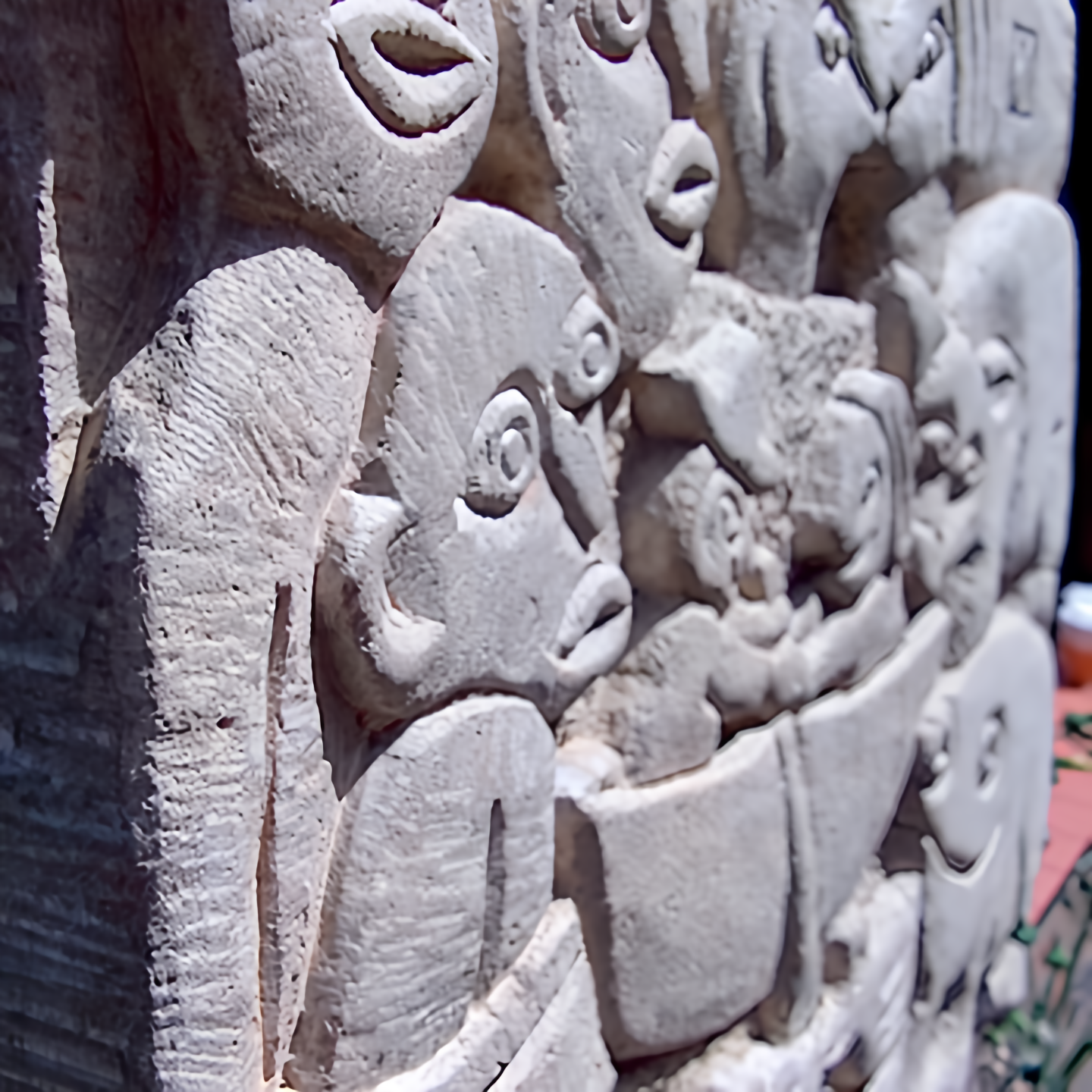}};
\spy on (0.8,0.8) in node [left] at (1,-0.6);
\draw (0, 1.2) node {SwinIR};
\end{tikzpicture}
\begin{tikzpicture}[spy using outlines={circle,yellow,magnification=3,size=1.5cm, connect spies}]
\node {\includegraphics[height=2cm]{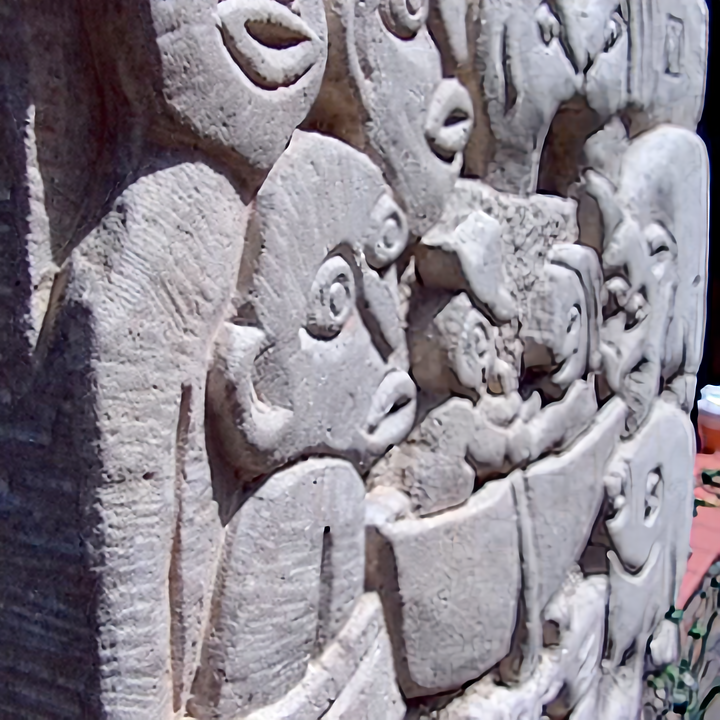}};
\spy on (0.8,0.8) in node [left] at (1,-0.6);
\draw (0, 1.2) node {BlindSR};
\end{tikzpicture}
\begin{tikzpicture}[spy using outlines={circle,yellow,magnification=3,size=1.5cm, connect spies}]
\node {\includegraphics[height=2cm]{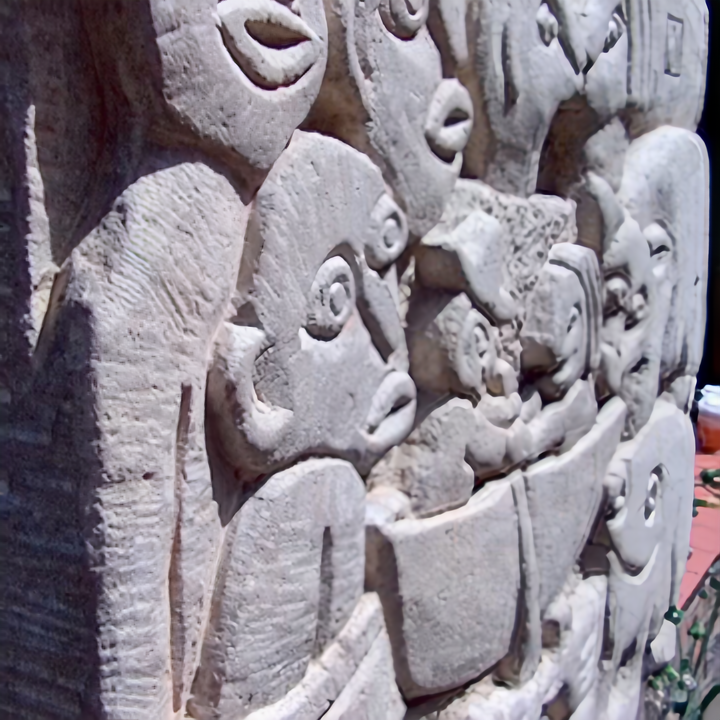}};
\spy on (0.8,0.8) in node [left] at (1,-0.6);
\draw (0, 1.2) node {Ours};
\end{tikzpicture}%
}
\caption{Super-resolution with scale factor s=2 on real-world defocused images}
\label{fig:blind_sr_defocus}
\end{figure}
The success of our method first relies on the fact that the kernel information does not need to be encoded to be fed to the model which allows good deblurring quality of very complex kernels. Also, the deblurring and super-resolution modules are fixed in our architecture which accounts for increased robustness to different blur types. It is worth pointing out that only a single version of our model was used for all the scale factors without the need to retraining. 

\subsection{Visual Results}
\begin{figure*}[t]
\centering
\resizebox{1.6\columnwidth}{!}{%
\begin{tabular}{c|ccccc}
 LR & SwinIR & BlindSR & USRNet & Ours & GT \\
\includegraphics[width=0.27\linewidth]{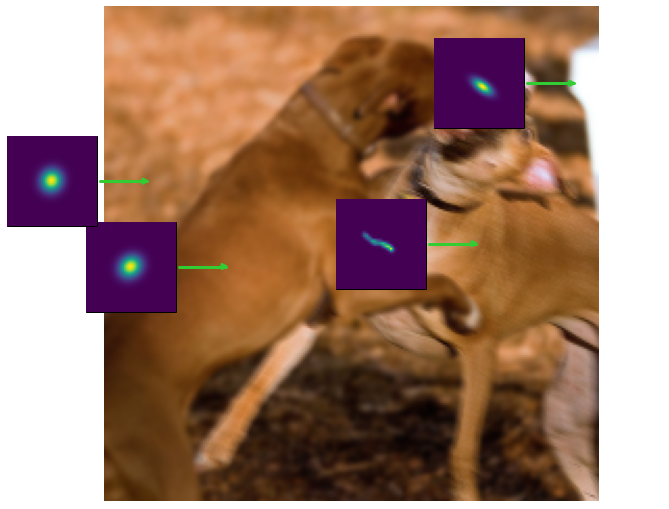} &
\includegraphics[width=0.2\linewidth]{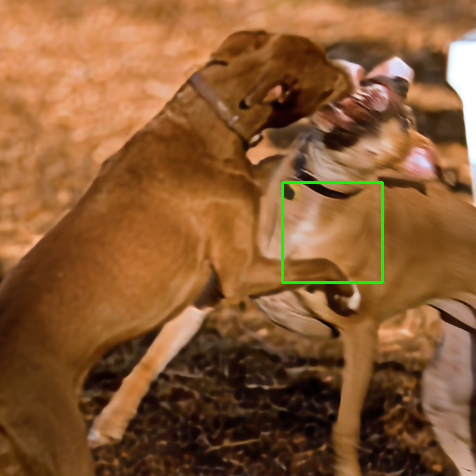} & 
\includegraphics[width=0.2\linewidth]{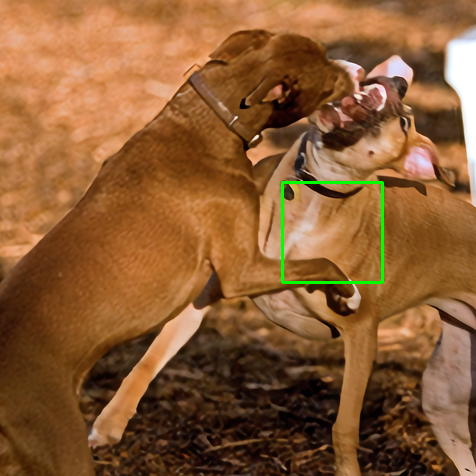} & 
\includegraphics[width=0.2\linewidth]{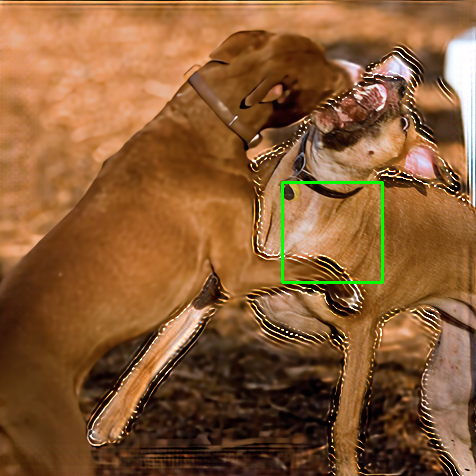} & 
\includegraphics[width=0.2\linewidth]{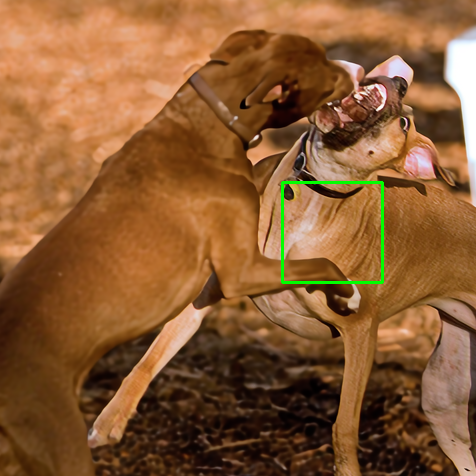} & 
\includegraphics[width=0.2\linewidth]{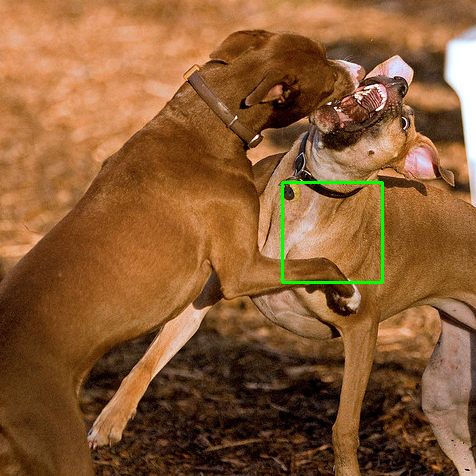} \\
\includegraphics[width=0.25\linewidth]{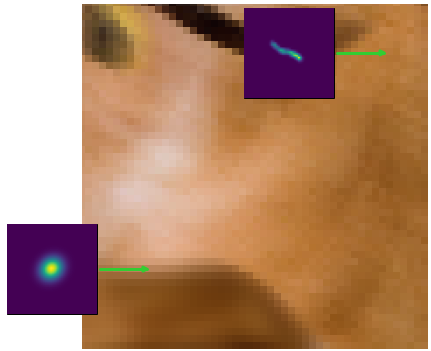} &
\includegraphics[width=0.2\linewidth]{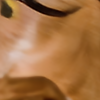} & 
\includegraphics[width=0.2\linewidth]{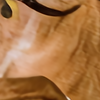} & 
\includegraphics[width=0.2\linewidth]{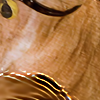} & 
\includegraphics[width=0.2\linewidth]{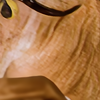} & 
\includegraphics[width=0.2\linewidth]{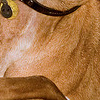} \\
\includegraphics[width=0.27\linewidth]{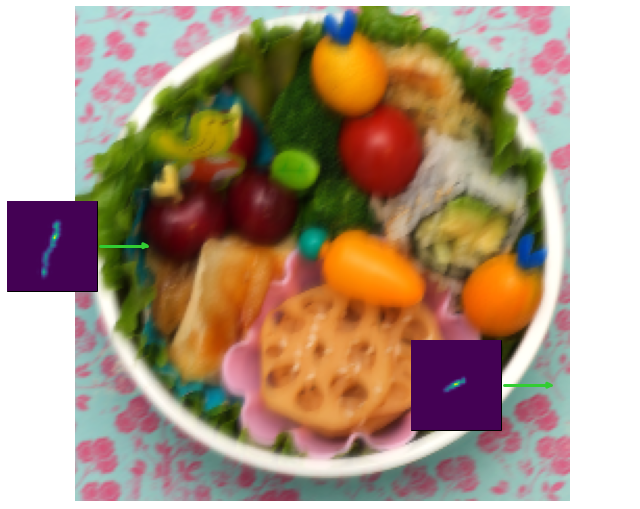} &
\includegraphics[width=0.2\linewidth]{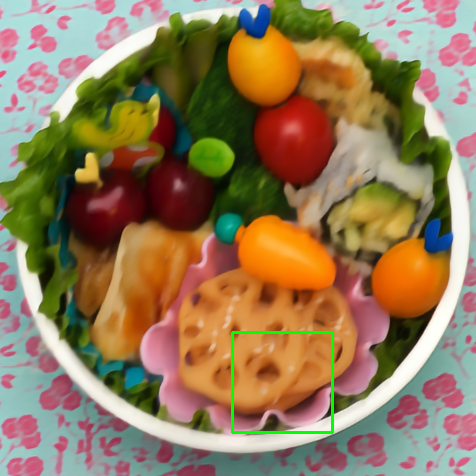} & 
\includegraphics[width=0.2\linewidth]{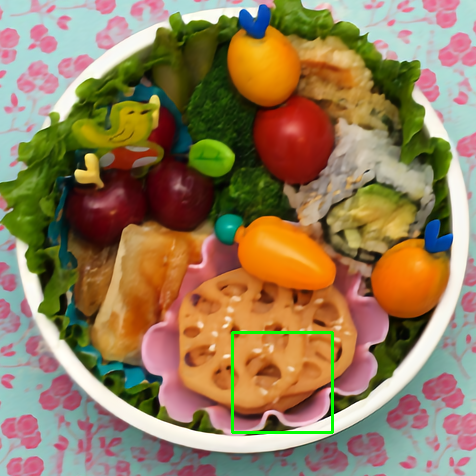} & 
\includegraphics[width=0.2\linewidth]{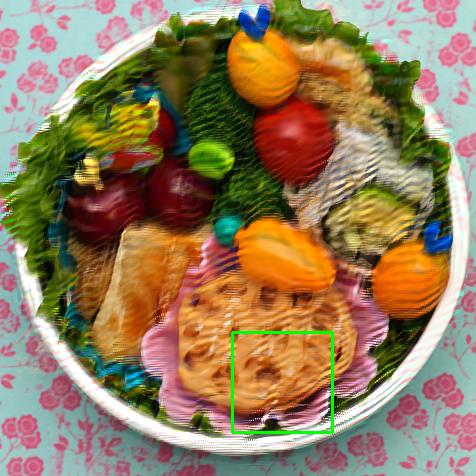} & 
\includegraphics[width=0.2\linewidth]{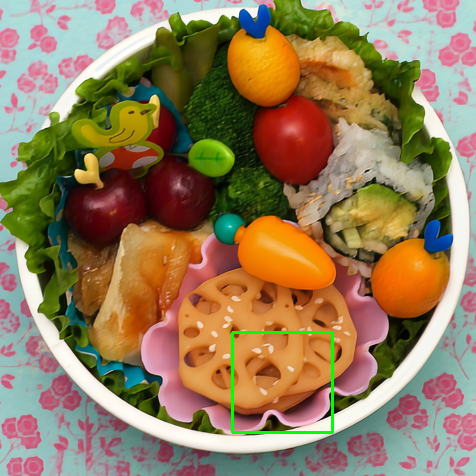} & 
\includegraphics[width=0.2\linewidth]{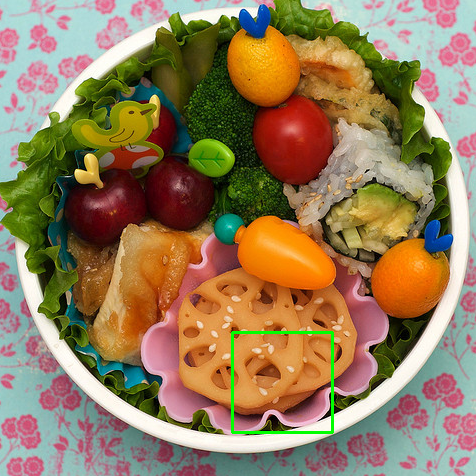} \\
\includegraphics[width=0.23\linewidth]{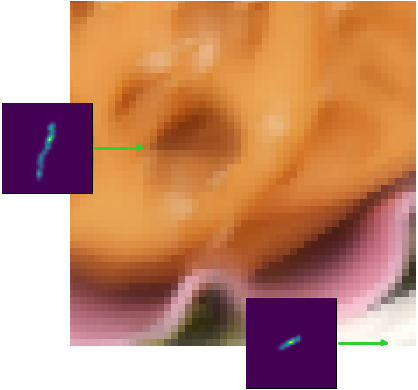} &
\includegraphics[width=0.2\linewidth]{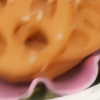} & 
\includegraphics[width=0.2\linewidth]{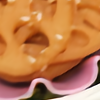} & 
\includegraphics[width=0.2\linewidth]{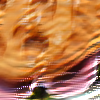} & 
\includegraphics[width=0.2\linewidth]{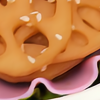} & 
\includegraphics[width=0.2\linewidth]{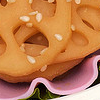} \\
\includegraphics[width=0.27\linewidth]{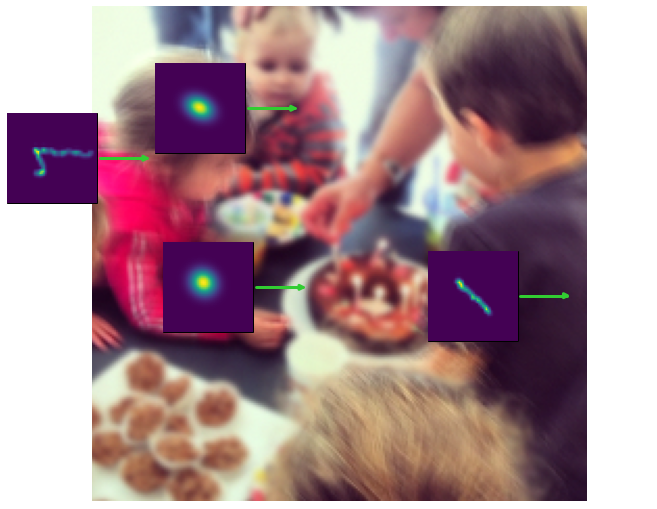} &
\includegraphics[width=0.2\linewidth]{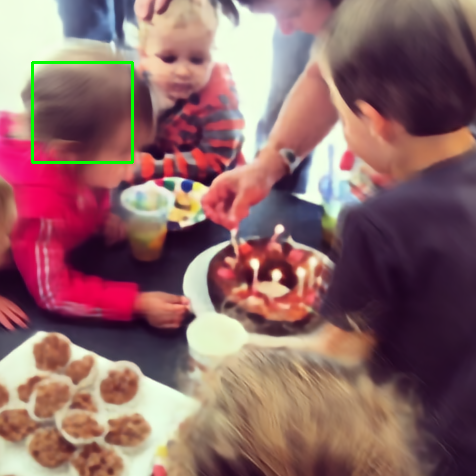} & 
\includegraphics[width=0.2\linewidth]{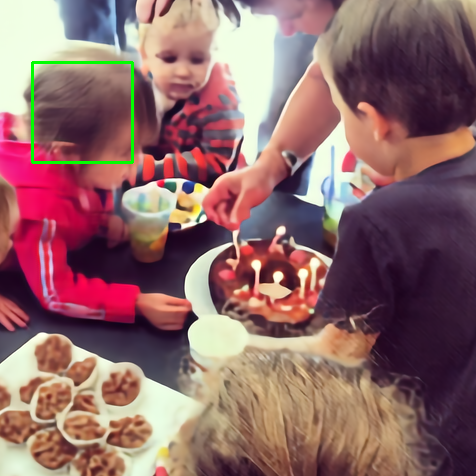} & 
\includegraphics[width=0.2\linewidth]{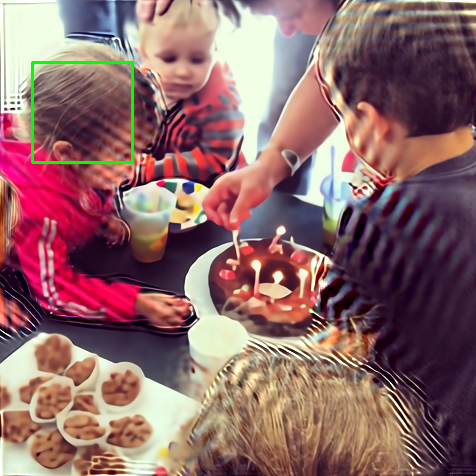} & 
\includegraphics[width=0.2\linewidth]{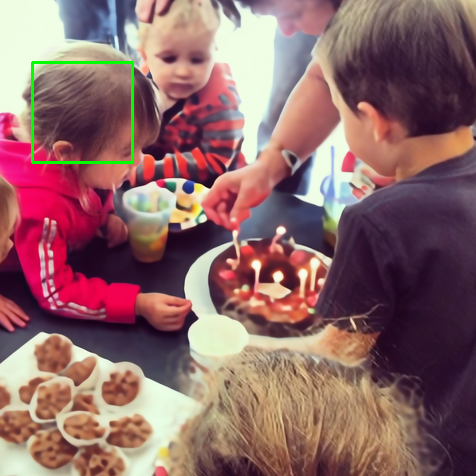} &
\includegraphics[width=0.2\linewidth]{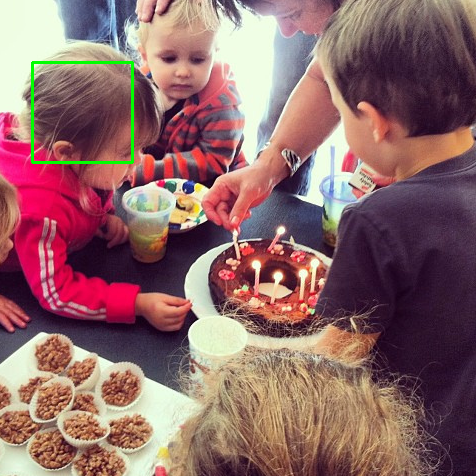} \\
\includegraphics[width=0.23\linewidth]{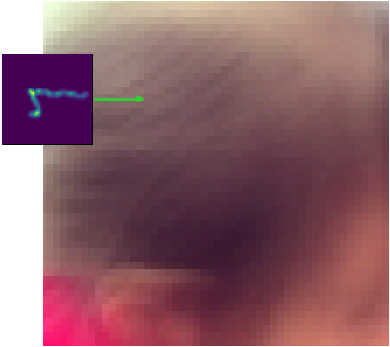} &
\includegraphics[width=0.2\linewidth]{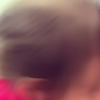} & 
\includegraphics[width=0.2\linewidth]{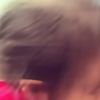} & 
\includegraphics[width=0.2\linewidth]{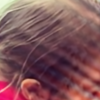} & 
\includegraphics[width=0.2\linewidth]{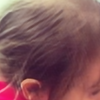} &
\includegraphics[width=0.2\linewidth]{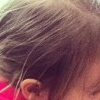}
\end{tabular}%
}
\caption{Visual comparison of the super-resolution performance of the models with a scale factor of 2. The different blur kernels are displayed in the LR images.}
\label{fig:visual_res}
\end{figure*}

Figure \ref{fig:visual_res} shows a visual comparisons of the different models on x2 super-resolution. We excluded RRDB and SFTMD since their performance are outmatched by SwinIR and BlindSR, respectively. We observe that SwinIR produces results that are still blurry. USRNet yields sharp results on the areas where the blur kernels are not mixed (\emph{i.e.}~when the area of a single degradation map is equal to one and all the other are equal to zeros), but introduces artefacts on the edges of the objects since the deblurring task is not linear. BlindSR super-resolves well the images. However, some areas remain blurry especially when there is motion blur. Finally, our model successfully produces a sharp super-resolved image without artefacts. We observe more texture details and sharper edges. 

\subsection{Real-world images}

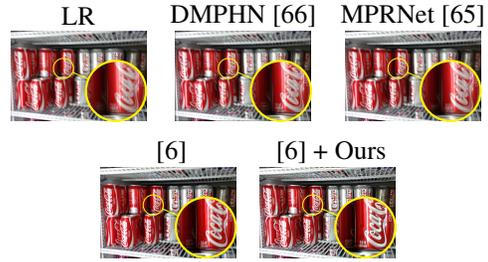
\begin{figure}[t]
\centering
\begin{tikzpicture}[spy using outlines={circle,yellow,magnification=3,size=0.8cm, connect spies}]
\node {\includegraphics[height=1.2cm]{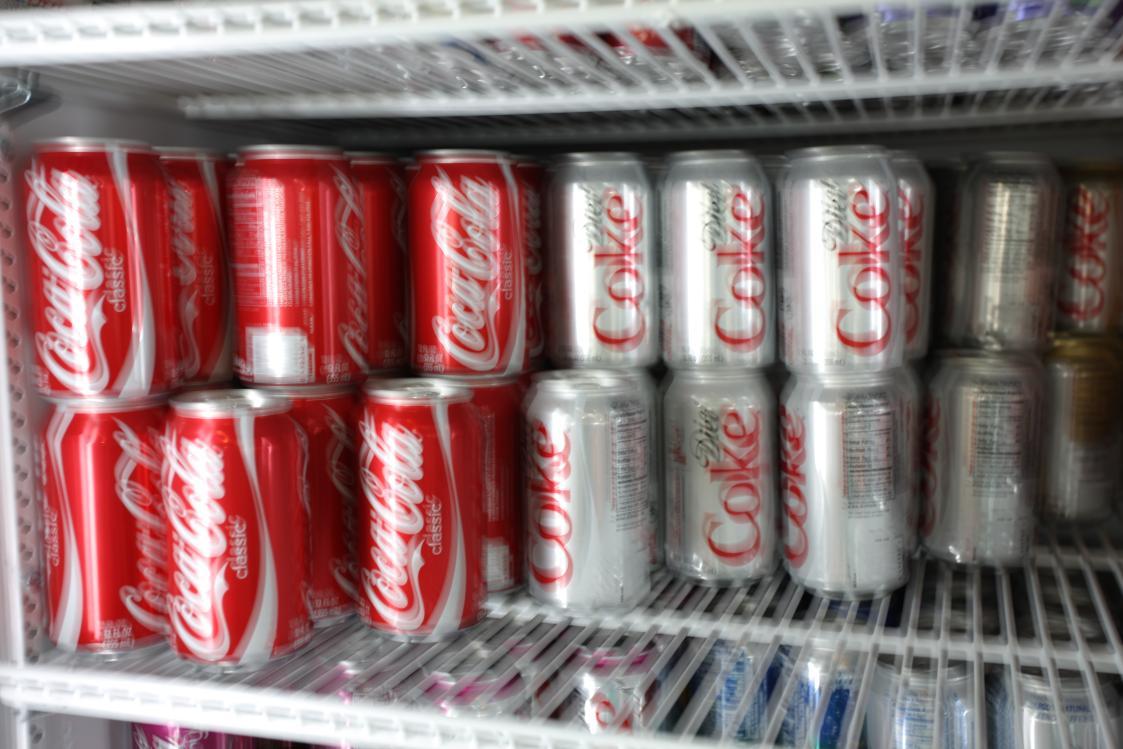}};
\spy on (-0.2,0.1) in node [left] at (0.9,-0.2);
\draw (0, 0.8) node {LR};
\end{tikzpicture}
\begin{tikzpicture}[spy using outlines={circle,yellow,magnification=3,size=0.8cm, connect spies}]
\node {\includegraphics[height=1.2cm]{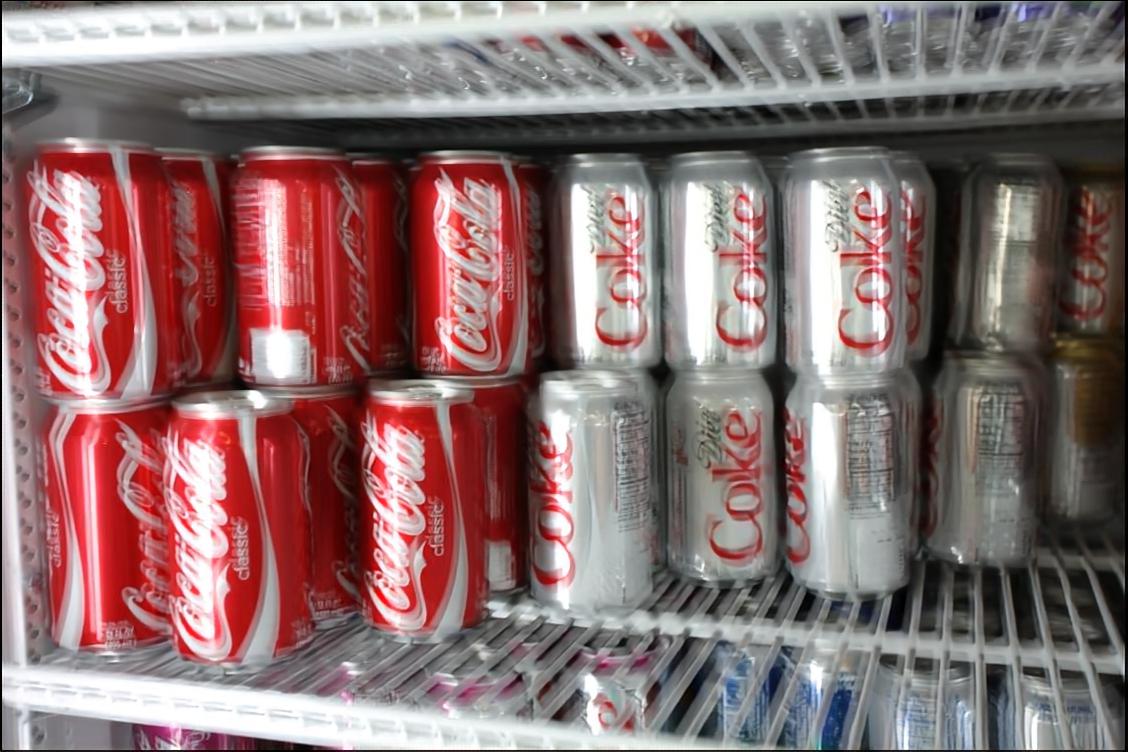}};
\spy on (-0.2,0.1) in node [left] at (0.9,-0.2);
\draw (0, 0.8) node {DMPHN~\cite{Zhang_2019_CVPR}};
\end{tikzpicture}
\begin{tikzpicture}[spy using outlines={circle,yellow,magnification=3,size=0.8cm, connect spies}]
\node {\includegraphics[height=1.2cm]{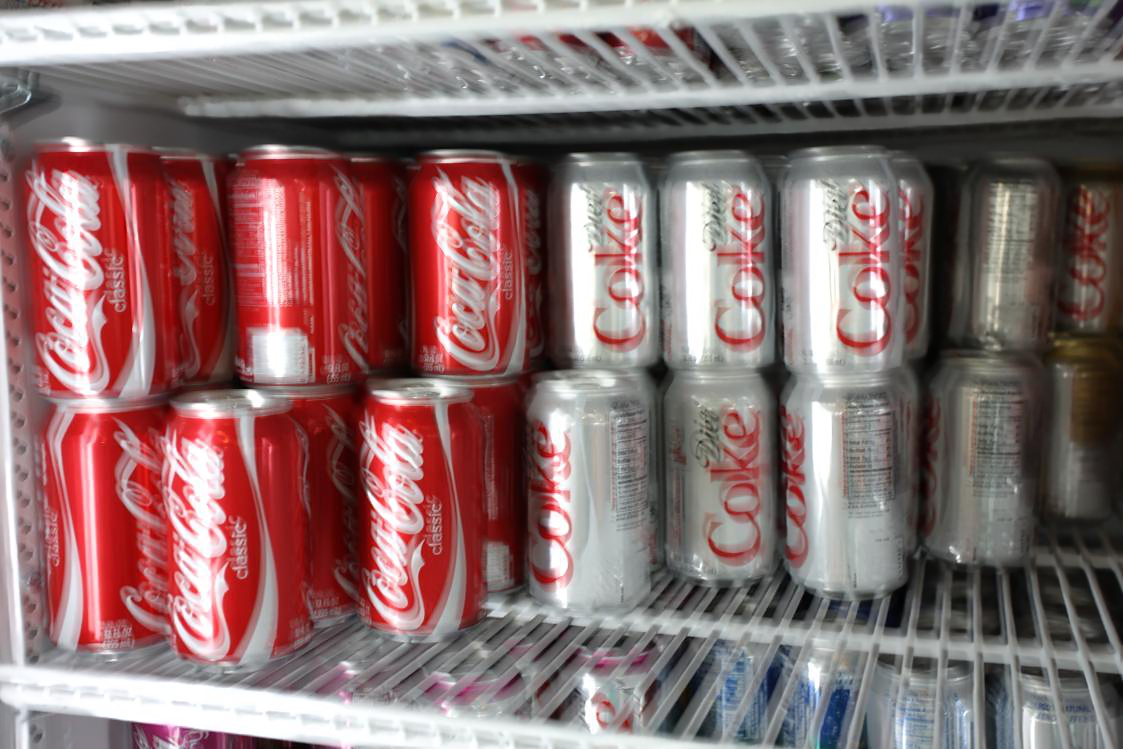}};
\spy on (-0.2,0.1) in node [left] at (0.9,-0.2);
\draw (0, 0.8) node {MPRNet~\cite{Zamir2021MPRNet}};
\end{tikzpicture}
\begin{tikzpicture}[spy using outlines={circle,yellow,magnification=3,size=0.8cm, connect spies}]
\node {\includegraphics[height=1.2cm]{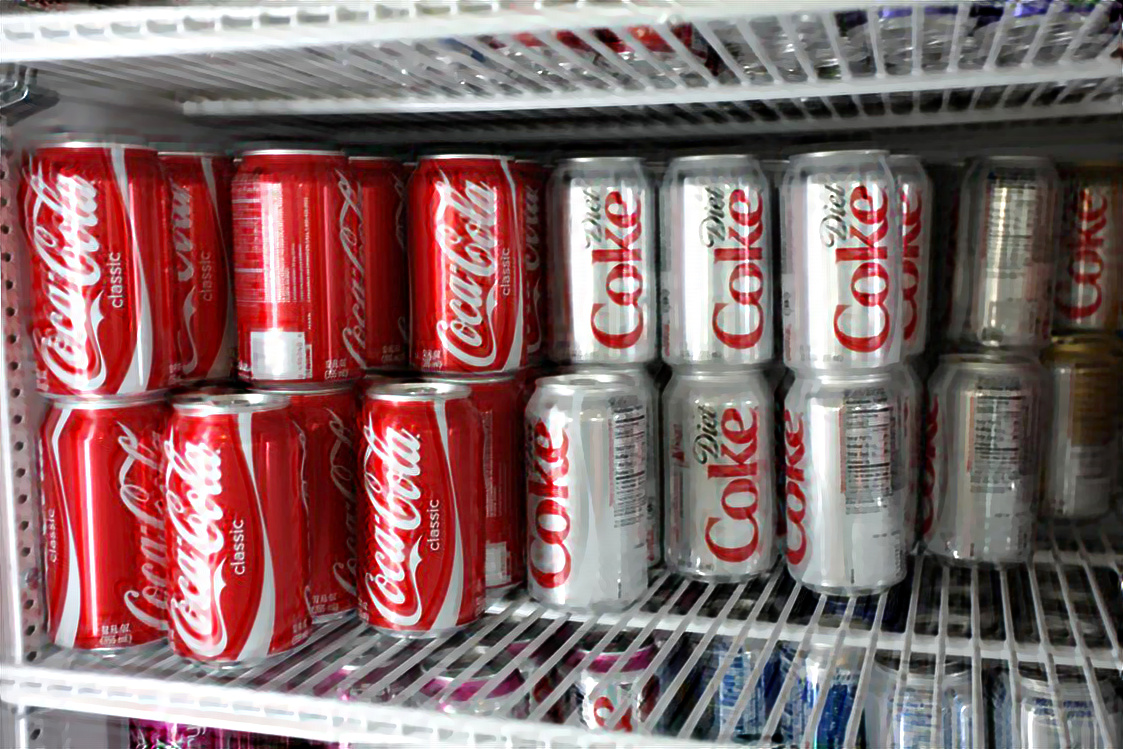}};
\spy on (-0.2,0.1) in node [left] at (0.9,-0.2);
\draw (0, 0.8) node {\cite{carbajal2021nonuniform}};
\end{tikzpicture}
\begin{tikzpicture}[spy using outlines={circle,yellow,magnification=3,size=0.8cm, connect spies}]
\node {\includegraphics[height=1.2cm]{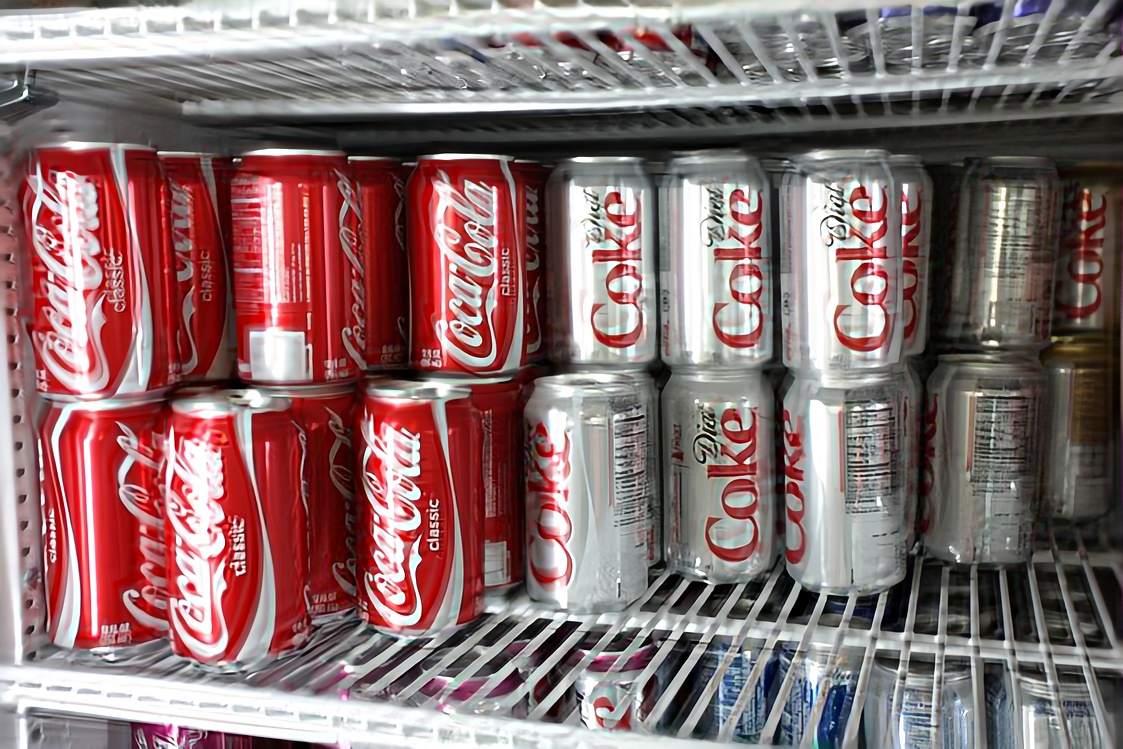}};
\spy on (-0.2,0.1) in node [left] at (0.9,-0.2);
\draw (0, 0.8) node {\cite{carbajal2021nonuniform} + Ours};
\end{tikzpicture}
\caption{Deblurring results on Lai~\cite{Lai-CVPR-2016} dataset}
\label{fig:blind_deblur}
\end{figure}
Testing our method on real-world images requires that we can access to the blur operator associated to the image. The performance of the super-resolution model strongly relies on the accuracy of the blur kernel. Figure~\ref{fig:blind_sr_defocus} shows SR results of the models on real-world defocus blur where the blur operator was estimated using camera properties~\cite{7401071} while Figure~\ref{fig:blind_deblur} displays deblurring results (scale factor $s=1$, no SR) where the blur operator was estimated using~\cite{carbajal2021nonuniform}.
In the first example, it can be observed that BlindSR tends to add over-sharpening artefacts in the image while performing super-resolution. SwinIR returns a cleaner image but that is still blurry. On the other hand, our model returns the sharpest result. These results also highlight the good generalization properties of our algorithm as the model was not trained on defocus blur kernels or smooth variations from a kernel to another at the image scale (the borders of the masks used in training are hard).
In deblurring, we compared our model to DMPHN~\cite{Zhang_2019_CVPR}, MPRNet~\cite{Zamir2021MPRNet} and the deblurring scheme employed in \cite{carbajal2021nonuniform}. We found that our model outperformed these methods both in terms of sharpness and deblurring artifacts.
More visual results can be found in the supplementary material and the webpage of the project.

\section{Conclusion \& Future Research}\label{sec:conclusion}
In this paper, we approach the problem of single-image super-resolution with spatially-varying blur. We propose a deep unfolding architecture that handles various blur kernels, scale factors, and noise levels. Our unfolding architecture derives from a deep plug-and-play algorithm based on the linearized ADMM splitting technique.
Our architecture inherits both from the flexibility of plug-and-play algorithms and from the speed and efficiency of learning-based methods through end-to-end training. Experimental results using the O’Leary blur model highlight the superiority of the proposed method in terms of performance and generalization. We also show that the model generalizes well to real-world data using existing kernel estimation methods.


\bibliographystyle{ieee_fullname}
\bibliography{egbib}

\begin{thebibliography}{10}\itemsep=-1pt

\bibitem{Almansa2004}
Andr{\'{e}}s Almansa, Sylvain Durand, and Bernard Roug{\'{e}}.
\newblock {Measuring and Improving Image Resolution by Adaptation of the
  Reciprocal Cell}.
\newblock {\em Journal of Mathematical Imaging and Vision}, 21(3):235--279, nov
  2004.

\bibitem{KernelGAN}
Sefi Bell-Kligler, Assaf Shocher, and Michal Irani.
\newblock Blind super-resolution kernel estimation using an internal-gan.
\newblock In {\em NeurIPS}, 2019.

\bibitem{Boyd2010a}
Stephen Boyd, Neal Parikh, Eric Chu, Borja Peleato, and Jonathan Eckstein.
\newblock {Distributed Optimization and Statistical Learning via the
  Alternating Direction Method of Multipliers}.
\newblock {\em Foundations and Trends{\textregistered} in Machine Learning},
  3(1):1--122, 2010.

\bibitem{dpnp_sr}
Alon Brifman, Yaniv Romano, and Michael Elad.
\newblock Turning a denoiser into a super-resolver using plug and play priors.
\newblock In {\em 2016 IEEE International Conference on Image Processing
  (ICIP)}, 2016.

\bibitem{dpnp_gauss}
Gregery~T. Buzzard, Stanley~H. Chan, Suhas Sreehari, and Charles~A. Bouman.
\newblock Plug-and-play unplugged: Optimization-free reconstruction using
  consensus equilibrium.
\newblock {\em SIAM Journal on Imaging Sciences}, 2018.

\bibitem{carbajal2021nonuniform}
Guillermo Carbajal, Patricia Vitoria, Mauricio Delbracio, Pablo Musé, and
  José Lezama.
\newblock Non-uniform blur kernel estimation via adaptive basis decomposition.
\newblock {\em arXiv:2102.01026}, 2021.

\bibitem{dpnp_theory_bounded}
Stanley~H. Chan, Xiran Wang, and Omar~A. Elgendy.
\newblock Plug-and-play admm for image restoration: Fixed-point convergence and
  applications.
\newblock {\em IEEE Transactions on Computational Imaging}, 2017.

\bibitem{BlindSR}
Victor Cornillère, Abdelaziz Djelouah, Wang Yifan, Olga Sorkine-Hornung, and
  Christopher Schroers.
\newblock Blind image super-resolution with spatially variant degradations.
\newblock In {\em ACM Transactions on Graphics}, 2019.

\bibitem{dpnp_bm3d}
Yehuda Dar, Alfred~M. Bruckstein, Michael Elad, and Raja Giryes.
\newblock Postprocessing of compressed images via sequential denoising.
\newblock {\em IEEE Transactions on Image Processing}, 2016.

\bibitem{ResUNet}
Foivos~I. Diakogiannis, François Waldner, Peter Caccetta, and Chen Wu.
\newblock Resunet-a: A deep learning framework for semantic segmentation of
  remotely sensed data.
\newblock {\em ISPRS Journal of Photogrammetry and Remote Sensing},
  162:94--114, 2020.

\bibitem{SRCNN}
Chao Dong, Chen~Change Loy, Kaiming He, and Xiaoou Tang.
\newblock Learning a deep convolutional network for image super-resolution.
\newblock In David Fleet, Tomas Pajdla, Bernt Schiele, and Tinne Tuytelaars,
  editors, {\em Proceedings of European Conference on Computer Vision (ECCV)},
  pages 184--199. Springer International Publishing, 2014.

\bibitem{7401071}
Laurent D’Andrès, Jordi Salvador, Axel Kochale, and Sabine Süsstrunk.
\newblock Non-parametric blur map regression for depth of field extension.
\newblock {\em IEEE Transactions on Image Processing}, 25(4):1660--1673, 2016.

\bibitem{eboli2020end2end}
Thomas Eboli, Jian Sun, and Jean Ponce.
\newblock End-to-end interpretable learning of non-blind image deblurring.
\newblock In {\em ECCV}, 2020.

\bibitem{DualSR}
Mohammad Emad, Maurice Peemen, and Henk Corporaal.
\newblock {DualSR: Zero-Shot Dual Learning for Real-World Super-Resolution}.
\newblock In {\em 2021 IEEE Winter Conference on Applications of Computer
  Vision (WACV)}, pages 1629--1638, 2021.

\bibitem{escande2018}
Paul Escande and Pierre Weiss.
\newblock {Accelerating l1-l2 deblurring using wavelet expansions of
  operators}.
\newblock {\em Journal of Computational and Applied Mathematics}, 343:373--396,
  dec 2018.

\bibitem{escande_sparse_2013}
Paul Escande, Pierre Weiss, and Fran{\c{c}}ois Malgouyres.
\newblock {Image restoration using sparse approximations of spatially varying
  blur operators in the wavelet domain}.
\newblock {\em Journal of Physics: Conference Series}, 464(1):012004, oct 2013.

\bibitem{escande_diag_2013}
Paul Escande, Pierre Weiss, and François Malgouyres.
\newblock Spatially varying blur recovery. diagonal approximations in the
  wavelet domain.
\newblock {\em ICPRAM}, 2013.

\bibitem{Esser2010}
Ernie Esser, Xiaoqun Zhang, and Tony~F. Chan.
\newblock A general framework for a class of first order primal-dual algorithms
  for convex optimization in imaging science.
\newblock In {\em Society for Industrial and Applied Mathematics (SIAM)}, 2010.

\bibitem{Delbracio2011psf}
Delbracio et al.
\newblock {The Non-parametric Sub-pixel Local Point Spread Function Estimation
  Is a Well Posed Problem}.
\newblock {\em IJCV}, jan 2012.

\bibitem{Ikoma2021dfd}
Ikoma et al.
\newblock Depth from defocus with learned optics for imaging and
  occlusion-aware depth estimation.
\newblock {\em ICCP}, 2021.

\bibitem{camera_shake}
Fabien Gavant, Laurent Alacoque, Antoine Dupret, and Dominique David.
\newblock A physiological camera shake model for image stabilization systems.
\newblock In {\em SENSORS, 2011 IEEE}, pages 1461--1464, 2011.

\bibitem{Gribonval2011}
R{\'{e}}mi Gribonval.
\newblock Should penalized least squares regression be interpreted as maximum a
  posteriori estimation?
\newblock {\em IEEE Transactions on Signal Processing}, 59(5):2405--2410, 2011.

\bibitem{IKC}
Jinjin Gu, Hannan Lu, Wangmeng Zuo, and Chao Dong.
\newblock Blind super-resolution with iterative kernel correction.
\newblock In {\em 2019 IEEE/CVF Conference on Computer Vision and Pattern
  Recognition (CVPR)}, pages 1604--1613, 2019.

\bibitem{dpnp_dualprimal}
Felix Heide, Markus Steinberger, Yun-Ta Tsai, Mushfiqur Rouf, Dawid Pajak,
  Dikpal Reddy, Orazio Gallo, Jing Liu, Wolfgang Heidrich, Karen Egiazarian,
  Jan Kautz, and Kari Pulli.
\newblock {FlexISP: A Flexible Camera Image Processing Framework}.
\newblock {\em ACM Transactions on Graphics}, 33, 2014.

\bibitem{dpnp_fista}
Ulugbek~S. Kamilov, Hassan Mansour, and Brendt Wohlberg.
\newblock A plug-and-play priors approach for solving nonlinear imaging inverse
  problems.
\newblock {\em IEEE Signal Processing Letters}, 2017.

\bibitem{vdsr}
Jiwon Kim, Jung~Kwon Lee, and Kyoung~Mu Lee.
\newblock Accurate image super-resolution using very deep convolutional
  networks.
\newblock In {\em 2016 IEEE Conference on Computer Vision and Pattern
  Recognition (CVPR)}, pages 1646--1654, 2016.

\bibitem{recur_sr}
Jiwon Kim, Jung~Kwon Lee, and Kyoung~Mu Lee.
\newblock Deeply-recursive convolutional network for image super-resolution.
\newblock In {\em 2016 IEEE Conference on Computer Vision and Pattern
  Recognition (CVPR)}, pages 1637--1645, 2016.

\bibitem{NIPS2009_3dd48ab3}
Dilip Krishnan and Rob Fergus.
\newblock Fast image deconvolution using hyper-laplacian priors.
\newblock In Y. Bengio, D. Schuurmans, J. Lafferty, C. Williams, and A.
  Culotta, editors, {\em Advances in Neural Information Processing Systems},
  volume~22. Curran Associates, Inc., 2009.

\bibitem{Lai-CVPR-2016}
Wei-Sheng Lai, Jia-Bin Huang, Zhe Hu, Narendra Ahuja, and Ming-Hsuan Yang.
\newblock A comparative study for single image blind deblurring.
\newblock In {\em IEEE Conferene on Computer Vision and Pattern Recognition},
  2016.

\bibitem{Laumont2022pnpsgd}
R{\'{e}}mi Laumont, Valentin de Bortoli, Andr{\'{e}}s Almansa, Julie Delon,
  Alain Durmus, and Marcelo Pereyra.
\newblock {On Maximum-a-Posteriori estimation with Plug {\&} Play priors and
  stochastic gradient descent}.
\newblock Technical report, MAP5, jan 2022.

\bibitem{l1deblur}
Anat Levin, Yair Weiss, Fredo Durand, and William~T. Freeman.
\newblock Understanding and evaluating blind deconvolution algorithms.
\newblock In {\em 2009 IEEE Conference on Computer Vision and Pattern
  Recognition}, pages 1964--1971, 2009.

\bibitem{SwinIR}
Jingyun Liang, Jiezhang Cao, Guolei Sun, Kai Zhang, Luc Van~Gool, and Radu
  Timofte.
\newblock Swinir: Image restoration using swin transformer.
\newblock In {\em 2021 IEEE/CVF International Conference on Computer Vision
  Workshops (ICCVW)}, pages 1833--1844, 2021.

\bibitem{MaNet}
Jingyun Liang, Guolei Sun, Kai Zhang, Luc Van~Gool, and Radu Timofte.
\newblock Mutual affine network for spatially variant kernel estimation in
  blind image super-resolution.
\newblock In {\em IEEE International Conference on Computer Vision (ICCV)},
  2021.

\bibitem{Lim2017}
Bee Lim, Sanghyun Son, Heewon Kim, Seungjun Nah, and Kyoung~Mu Lee.
\newblock Enhanced deep residual networks for single image super-resolution.
\newblock In {\em 2017 IEEE Conference on Computer Vision and Pattern
  Recognition Workshops (CVPRW)}, pages 1132--1140, 2017.

\bibitem{coco}
Tsung-Yi Lin, Michael Maire, Serge Belongie, Lubomir Bourdev, Ross Girshick,
  James Hays, Pietro Perona, Deva Ramanan, C.~Lawrence Zitnick, and Piotr
  Doll\'ar.
\newblock {Microsoft COCO: Common Objects in Context}.
\newblock In {\em (ECCV) European Conference on Computer Vision}, 2015.

\bibitem{zhouchen2011}
Zhouchen Lin, Risheng Liu, and Zhixun Su.
\newblock Linearized alternating direction method with adaptive penalty for
  low-rank representation.
\newblock In {\em Advances in Neural Information Processing Systems}. Curran
  Associates, Inc., 2011.

\bibitem{Liu2019}
Qinghua Liu, Xinyue Shen, and Yuantao Gu.
\newblock {Linearized ADMM for Nonconvex Nonsmooth Optimization With
  Convergence Analysis}.
\newblock {\em IEEE Access}, 7:76131--76144, 2019.

\bibitem{swin}
Ze Liu, Yutong Lin, Yue Cao, Han Hu, Yixuan Wei, Zheng Zhang, Stephen Lin, and
  Baining Guo.
\newblock Swin transformer: Hierarchical vision transformer using shifted
  windows.
\newblock In {\em International Conference on Computer Vision (ICCV)}, 2021.

\bibitem{Malgouyres_2002}
F Malgouyres and F Guichard.
\newblock {Edge Direction Preserving Image Zooming: A Mathematical and
  Numerical Analysis}.
\newblock {\em SIAM Journal on Numerical Analysis}, 39(1):1--37, jan 2001.

\bibitem{Meinhardt2017}
Tim Meinhardt, Michael Moeller, Caner Hazirbas, and Daniel Cremers.
\newblock {Learning Proximal Operators: Using Denoising Networks for
  Regularizing Inverse Imaging Problems}.
\newblock In {\em (ICCV) International Conference on Computer Vision}, pages
  1799--1808. IEEE, oct 2017.

\bibitem{eff}
Hirsch Michael, Sra Suvrit, Schölkopf Bernhard, and Harmeling Stefan.
\newblock Efficient filter flow for space-variant multiframe blind
  deconvolution.
\newblock In {\em IEEE Computer Society Conference on Computer Vision and
  Pattern Recognition}, 2010.

\bibitem{Michaeli2013}
Tomer Michaeli and Michal Irani.
\newblock {Nonparametric Blind Super-resolution}.
\newblock In {\em (ICCV) International Conference on Computer Vision}, pages
  945--952. IEEE, dec 2013.

\bibitem{Michaeli2014}
Tomer Michaeli and Michal Irani.
\newblock {Blind Deblurring Using Internal Patch Recurrence}.
\newblock In {\em (ECCV) European Conference on Computer Vision}, volume 8691
  LNCS, pages 783--798. Springer, 2014.

\bibitem{milanfar2011SRbook}
Peyman Milanfar, editor.
\newblock {\em {Super-Resolution Imaging}}.
\newblock CRC Press, dec 2011.

\bibitem{oleary}
James~G. Nagy and Dianne~P. O'Leary.
\newblock Restoring images degraded by spatially variant blur.
\newblock {\em SIAM Journal on Scientific Computing}, 19(4):1063--1082, 1998.

\bibitem{stoch_ADMM}
Hua Ouyang, Niao He, and Alexander Gray.
\newblock {Stochastic ADMM for Nonsmooth Optimization}.
\newblock {\em arXiv:1211.0632}, nov 2012.

\bibitem{prox_algo}
Neal Parikh and Stephen Boyd.
\newblock {Proximal Algorithms}.
\newblock {\em Foundations and Trends{\textregistered} in Optimization},
  1(3):127--239, 2014.

\bibitem{Protter_2009}
Matan Protter, Michael Elad, Hiroyuki Takeda, and Peyman Milanfar.
\newblock {Generalizing the nonlocal-means to super-resolution reconstruction.}
\newblock In {\em IEEE Transactions on Image Processing}, volume~18, pages
  36--51, jan 2009.

\bibitem{Romano2017}
Yaniv Romano, John Isidoro, and Peyman Milanfar.
\newblock {RAISR: Rapid and Accurate Image Super Resolution}.
\newblock {\em IEEE Transactions on Computational Imaging}, 3(1):110--125, jun
  2016.

\bibitem{l2deblur}
Leonid~I. Rudin, Stanley Osher, and Emad Fatemi.
\newblock Nonlinear total variation based noise removal algorithms.
\newblock {\em Physica D: Nonlinear Phenomena}, 60(1):259--268, 1992.

\bibitem{pmlr-v97-ryu19a}
Ernest Ryu, Jialin Liu, Sicheng Wang, Xiaohan Chen, Zhangyang Wang, and Wotao
  Yin.
\newblock Plug-and-play methods provably converge with properly trained
  denoisers.
\newblock In Kamalika Chaudhuri and Ruslan Salakhutdinov, editors, {\em
  Proceedings of the 36th International Conference on Machine Learning},
  volume~97 of {\em Proceedings of Machine Learning Research}, pages
  5546--5557, Long Beach, California, USA, 09--15 Jun 2019. PMLR.

\bibitem{Saharia2021}
Chitwan Saharia, Jonathan Ho, William Chan, Tim Salimans, David~J. Fleet, and
  Mohammad Norouzi.
\newblock {Image Super-Resolution via Iterative Refinement}.
\newblock {\em arXiv:2104.07636}, 2021.

\bibitem{7274732}
Christian~J. Schuler, Michael Hirsch, Stefan Harmeling, and Bernhard
  Schölkopf.
\newblock Learning to deblur.
\newblock {\em IEEE Transactions on Pattern Analysis and Machine Intelligence},
  38(7):1439--1451, 2016.

\bibitem{ZSSR}
Assaf Shocher, Nadav Cohen, and Michal Irani.
\newblock Zero-shot super-resolution using deep internal learning.
\newblock In {\em 2018 IEEE/CVF Conference on Computer Vision and Pattern
  Recognition}, pages 3118--3126, 2018.

\bibitem{vsorel2017towards}
Michal {\v{S}}orel, Filip {\v{S}}roubek, and Jan Flusser.
\newblock {Towards Super-Resolution in the Presence of Spatially Varying Blur}.
\newblock In {\em Super-Resolution Imaging}, chapter~7, pages 187--218. CRC
  Press, dec 2017.

\bibitem{persist_sr}
Ying Tai, Jian Yang, Xiaoming Liu, and Chunyan Xu.
\newblock Memnet: A persistent memory network for image restoration.
\newblock In {\em 2017 IEEE International Conference on Computer Vision
  (ICCV)}, pages 4549--4557, 2017.

\bibitem{8578951}
Xin Tao, Hongyun Gao, Xiaoyong Shen, Jue Wang, and Jiaya Jia.
\newblock Scale-recurrent network for deep image deblurring.
\newblock In {\em 2018 IEEE/CVF Conference on Computer Vision and Pattern
  Recognition}, pages 8174--8182, 2018.

\bibitem{dpnp_cnn_2}
Tom Tirer and Raja Giryes.
\newblock Image restoration by iterative denoising and backward projections.
\newblock {\em IEEE Transactions on Image Processing}, 2019.

\bibitem{venkatakrishnan2013}
Singanallur~V. Venkatakrishnan, Charles~A. Bouman, and Brendt Wohlberg.
\newblock Plug-and-play priors for model based reconstruction.
\newblock {\em IEEE Global Conference on Signal and Information Processing},
  2013.

\bibitem{dpnp_deblur}
Xiran Wang and Stanley~H. Chan.
\newblock Parameter-free plug-and-play admm for image restoration.
\newblock In {\em 2017 IEEE International Conference on Acoustics, Speech and
  Signal Processing (ICASSP)}, 2017.

\bibitem{sft}
Xintao Wang, Ke Yu, Chao Dong, and Chen Change~Loy.
\newblock Recovering realistic texture in image super-resolution by deep
  spatial feature transform.
\newblock In {\em 2018 IEEE/CVF Conference on Computer Vision and Pattern
  Recognition}, pages 606--615, 2018.

\bibitem{esrgan}
Xintao Wang, Ke Yu, Shixiang Wu, Jinjin Gu, Yihao Liu, Chao Dong, Yu Qiao, and
  Chen~Change Loy.
\newblock Esrgan: Enhanced super-resolution generative adversarial networks.
\newblock In {\em The European Conference on Computer Vision Workshops
  (ECCVW)}, 2018.

\bibitem{whyte10}
Oliver Whyte, Josef Sivic, Andrew Zisserman, and Jean Ponce.
\newblock Non-uniform deblurring for shaken images.
\newblock In {\em 2010 IEEE Computer Society Conference on Computer Vision and
  Pattern Recognition}, pages 491--498, 2010.

\bibitem{atten_sr}
Zhang Yulun, Li Kunpeng, Li Kai, Wang Lichen, Zhong Bineng, and Fu Yun.
\newblock Image super-resolution using very deep residual channel attention
  networks.
\newblock In {\em Proceedings of European Conference on Computer Vision
  (ECCV)}, 2018.

\bibitem{Zamir2021MPRNet}
Syed~Waqas Zamir, Aditya Arora, Salman Khan, Munawar Hayat, Fahad~Shahbaz Khan,
  Ming-Hsuan Yang, and Ling Shao.
\newblock Multi-stage progressive image restoration.
\newblock In {\em CVPR}, 2021.

\bibitem{Zhang_2019_CVPR}
Hongguang Zhang, Yuchao Dai, Hongdong Li, and Piotr Koniusz.
\newblock Deep stacked hierarchical multi-patch network for image deblurring.
\newblock In {\em The IEEE Conference on Computer Vision and Pattern
  Recognition (CVPR)}, June 2019.

\bibitem{zhang2021plug}
Kai Zhang, Yawei Li, Wangmeng Zuo, Lei Zhang, Luc Van~Gool, and Radu Timofte.
\newblock Plug-and-play image restoration with deep denoiser prior.
\newblock {\em IEEE Transactions on Pattern Analysis and Machine Intelligence},
  2021.

\bibitem{BSRGAN}
Kai Zhang, Jingyun Liang, Luc Van~Gool, and Radu Timofte.
\newblock Designing a practical degradation model for deep blind image
  super-resolution.
\newblock In {\em IEEE International Conference on Computer Vision}, pages
  4791--4800, 2021.

\bibitem{DUN}
Kai Zhang, Luc Van~Gool, and Radu Timofte.
\newblock Deep unfolding network for image super-resolution.
\newblock In {\em 2020 IEEE/CVF Conference on Computer Vision and Pattern
  Recognition (CVPR)}, pages 3214--3223, 2020.

\bibitem{zhang2018learning}
Kai Zhang, Wangmeng Zuo, and Lei Zhang.
\newblock Learning a single convolutional super-resolution network for multiple
  degradations.
\newblock In {\em 2018 IEEE/CVF Conference on Computer Vision and Pattern
  Recognition}, pages 3262--3271, 2018.

\bibitem{Zhang2011}
Xiaoqun Zhang, Martin Burger, and Stanley Osher.
\newblock A unified primal-dual algorithm framework based on bregman iteration.
\newblock In {\em Journal of Scientific Computing}, 2011.

\bibitem{FastSR}
Ningning Zhao, Qi Wei, Adrian Basarab, Nicolas Dobigeon, Denis Kouam{\'e}, and
  Jean-Yves Tourneret.
\newblock {Fast Single Image Super-Resolution Using a New Analytical Solution
  for l2--l2 Problems}.
\newblock {\em {IEEE Transactions on Image Processing}}, vol. 25(n{\textdegree}
  8):pp. 3683--3697, Aug. 2016.

\bibitem{6583957}
Xiang Zhu, Scott Cohen, Stephen Schiller, and Peyman Milanfar.
\newblock Estimating spatially varying defocus blur from a single image.
\newblock {\em IEEE Transactions on Image Processing}, 22(12):4879--4891, 2013.

\end{thebibliography}

\end{document}